%% file: main.tex
\documentclass[10pt,twocolumn]{article}

\usepackage[margin=0.75in]{geometry}
\usepackage[T1]{fontenc}
\usepackage[utf8]{inputenc}
\usepackage{lmodern}
\usepackage{microtype}
\usepackage{amsmath,amssymb}
\usepackage{booktabs}
\usepackage{multirow}
\usepackage{array}
\usepackage{tabularx}
\usepackage{graphicx}
\usepackage{xurl}
\usepackage{xcolor}
\usepackage{caption}
\usepackage{subcaption}
\usepackage{enumitem}
\usepackage{listings}
\usepackage{algorithm}
\usepackage{algpseudocode}
\usepackage{tikz}
\usetikzlibrary{positioning,arrows.meta,calc}
\usepackage[numbers,sort&compress]{natbib}
\usepackage[colorlinks=true,linkcolor=blue!50!black,citecolor=blue!50!black,urlcolor=blue!50!black]{hyperref}
\usepackage[capitalise,noabbrev]{cleveref}

\makeatletter
\def\theHALG@line{\thealgorithm.\arabic{ALG@line}}  
\makeatother
\setlist{itemsep=1pt,topsep=3pt}

\lstdefinestyle{py}{language=Python,basicstyle=\ttfamily\footnotesize,keywordstyle=\color{blue!60!black},
  commentstyle=\color{gray},stringstyle=\color{teal!70!black},showstringspaces=false,frame=single,
  framerule=0.3pt,rulecolor=\color{gray!60},xleftmargin=4pt,xrightmargin=4pt,columns=fullflexible,
  breaklines=true}
\newcommand{\lb}{SP}
\newcommand{\nvarc}{NVARC~4B}
\newcommand{\qbig}{Qwen3.8-27B}
\newcommand{\qflash}{Qwen3.8-Flash-Next}
\newcommand{\rftv}[1]{RFT~v#1}

\title{Rules, Traces and Routing Statistics:\\
Training and Calibration Data for Open Models on ARC-AGI-2}

\author{Nguyen Minh Thang$^{*}$, Trinh Thi Phuong\\
\small VNU University of Science, 334 Nguyen Trai, Thanh Xuan, Hanoi, Vietnam\\
\small\href{mailto:21025092@vnu.edu.vn}{\texttt{21025092@vnu.edu.vn}}, \href{mailto:phiutrinh@gmail.com}{\texttt{phiutrinh@gmail.com}}}

\date{November 2026}

\begin{document}
\maketitle

\input{sec/abstract}
\input{sec/intro}
\input{sec/background}
\input{sec/rules}
\input{sec/nvarc}
\input{sec/qwen}
\input{sec/flashnext}
\input{sec/results}
\input{sec/discussion}
\input{sec/conclusion}

\bibliographystyle{unsrtnat}
\bibliography{refs}

\appendix
\input{sec/appendix}

\end{document}

%% file: sec/abstract.tex
\begin{abstract}
We study which training and calibration data help three open models solve ARC-AGI-2 tasks when the whole solver runs
offline on four L4 GPUs for at most twelve hours. We compare data offline on public tasks and report 189 runs on the 120
hidden semi-private tasks. For the NVARC Qwen3-4B grid model, we built a corpus of 12,000 executable rules in a typed
language and 95,993 tasks; a rule is admitted only if its behaviour signature on a fixed probe set differs from those of
all admitted rules, and a task only if its pairs refute a shortlist of nearest competing rules and defeat 12 shortcut
solvers. Continued fine-tuning on 1,920 real task records, run on all four GPUs with revised scheduling, raised the
semi-private score from 28.96 (mean of the released model) to 33.06 (mean of five runs, best 34.86); no checkpoint trained
with synthetic tasks, including one trained on all 95,993 corpus tasks, improved on the real-task model. For the
Qwen3.8-27B reasoning model, we fine-tuned on 1,036 of its own reasoning traces whose final answers match the known
outputs, and compared this set with sets that add traces of other models and synthetic tasks. With an inference procedure
built around the model (the thinking cap enforced inside the serving engine, a Python program written after each answer
and checked on the demonstrations, a long first attempt for every input), the score rose from 26.81 to 44.03, the level
that the ARC Prize Foundation measured for the base model without a time limit (42.36); most of the gain came from the
inference procedure. For the Qwen3.8-Flash-Next mixture-of-experts model (125B parameters, 6B active), GPU memory limited
concurrency. Pruning from 512 to 384 experts per layer by router-weighted activation saliency grew the key-value cache from
0.55M to 1.39M tokens; a 296-expert list adapted with routing statistics from ARC training tasks and with experts kept for
program tokens, together with an FP8 indexer cache, reached 1.96M. With a tool that checks programs, the model scored
56.67, our best result.
\end{abstract}

%% file: sec/intro.tex
\section{Introduction}
\label{sec:intro}

ARC-AGI-2 asks a solver to infer a grid transformation from a few demonstration pairs and apply it to new inputs
\citep{chollet2019measure,chollet2025arcagi2}. Its hidden tasks are scored on a fixed computer: the solver runs without
internet access on four NVIDIA L4 GPUs for at most twelve hours, over 240 hidden tasks \citep{chollet2025arcagi2}. Under
this budget, the best private-set score reported for 2025 was 24.03\%, by NVARC \citep{chollet2026arcprize2025}. A solver
in this setting must be small or fast. This paper is about the data that a model learns from, or is calibrated on,
before the run, and about how much of the measured gains such data explain next to changes in serving and scheduling.

We worked with three open models that fit the budget: the NVARC Qwen3-4B grid model, which outputs the answer grid and is
trained again on each task at test time \citep{sorokin2025nvarc,franzen2024architects}; the dense \qbig{} reasoning model
\citep{qwen2026qwen38card}, which thinks for tens of thousands of tokens before it answers; and the \qflash{} reasoning model
\citep{qiu2026qwen38next}, a mixture of experts with 6B of its 125B parameters active per token. For each model we compared
training or calibration data, offline on public tasks and, for the main configurations, on the 120 hidden semi-private
tasks (189 runs in total).
Our contributions are:
\begin{itemize}
\item \emph{A corpus of 12,000 signature-distinct rules with identifiability checks} (\cref{sec:rules}). Each rule is a
typed program admitted only if its behaviour signature (on 220 probe grids and 352 typed slots) differs from that of every
other rule, and each of the 95,993 tasks only if, when it was generated, its pairs refuted a shortlist of nearest
competing rules and defeated 12 shortcut solvers. The corpus, with a set of 1,000 tasks that covers every
grid size from $1\times1$ to $30\times30$, supplied the synthetic data of the experiments below, including 2,231 synthetic
tasks in the fine-tuning experiments of \qbig{} (\cref{sec:qwen-synth}).
\item \emph{Real and synthetic training data for a grid model} (\cref{sec:nvarc}). We compared continued fine-tuning of
\nvarc{} on 1,920 real tasks with two FORGE50 recipes, a shape-balanced curriculum and checkpoints trained on the rule
corpus (\cref{tab:data-summary}). Fine-tuning on the real tasks raised the mean semi-private score from 28.96 to 31.72, and
running the fine-tuned model on all four GPUs with revised scheduling raised it to 33.06 (best 34.86); no checkpoint
trained with synthetic tasks, alone or next to the real-task model, raised the mean.
\item \emph{Answer-verified self-traces for a reasoning model} (\cref{sec:qwen}). We trained \qbig{} on 1,036 self-generated
traces whose answers match the known outputs, selected by closure, loop and give-up filters, and compared this set with
later sets that add second-pass or shorter traces, rule explanations, program targets, harder samples, traces of other
models and synthetic tasks (\cref{tab:qwen-data}). Around the fine-tuned model we
built an inference procedure: the thinking cap is enforced inside the serving engine, a Python program written after each
answer is checked on the demonstrations, and the long first attempt of every input is started before any second
attempt. The score rose
from 26.81 to 44.03, the level that the ARC Prize Foundation measured for the base model without a time limit; most of
the gain came from the inference procedure, and offline validation of the deployed schedule matched the semi-private
score.
\item \emph{Calibration data for expert pruning} (\cref{sec:flash}). We pruned \qflash{} from 512 to 384 experts per layer
by router-weighted activation saliency \citep{lasby2026reap} and then to 296, exchanging the least-used experts by routing
statistics from ARC training tasks and keeping experts for program tokens. The coding-calibrated saliency lists and the
ARC routing-frequency lists selected substantially different experts, and a list chosen by frequency alone lost accuracy.
Pruning, an FP8 indexer cache and a larger cache reservation grew the key-value cache 3.6 times, and \qflash{} reached
56.67, our best score.
\item \emph{An analysis of the fixed budget} (\cref{sec:discussion}): tokens per input constrain the score, answers forced
at the token cap are rarely right, and a program checked on the demonstrations gives a second answer of high precision.
\item \emph{Unsuccessful and incomplete experiments} (\cref{app:hparams,app:failed}). We report the hyper-parameters and
the variants that did not help with their measured results, including 16 expert lists and checkpoints of \qflash{} that
we rejected (\cref{tab:flash-failed}).
\end{itemize}
All scores are semi-private scores (\lb{}) of single twelve-hour runs unless noted; repeated runs of identical \nvarc{}
configurations vary with a standard deviation of 1.94 points (\cref{sec:results}), and we have too few repeats to estimate
the variation of the other two models.

%% file: sec/background.tex
\section{Background}
\label{sec:background}

\subsection{ARC-AGI-2 and its evaluation protocol}

An ARC task \citep{chollet2019measure} consists of a few demonstration pairs of coloured grids (up to $30\times30$ cells,
ten colours) and one or more test inputs; the solver must produce the exact output grid of each test input. ARC-AGI-2
\citep{chollet2025arcagi2} provides 1,000 training tasks and 120 public evaluation tasks with 172 test inputs, and keeps
240 further tasks hidden: 120 semi-private and 120 private tasks. A solver may give two attempts per test input, and the
score is the mean over tasks of the fraction of test inputs answered correctly. The hidden tasks are scored by running the
solver without internet access on four NVIDIA L4 GPUs (24~GB each) for at most 12 hours \citep{chollet2025arcagi2}. One
run covers the 120 semi-private and the 120 private tasks together \citep{chollet2025arcagi2}; at the 1.43 test inputs per
task of the public evaluation set this is about 340 inputs. Served on the four GPUs with an FP8 key-value cache and
multi-token prediction, \qbig{} generates about 490 tokens per second, or about 21M tokens in twelve hours and 62K tokens
per input, so a long first attempt fits for every input but a second attempt of the same length does not. We write K
and M for thousands and millions of tokens; thinking caps are round thousands (a 56K cap is 56,000 tokens). We report the score on
the 120 semi-private tasks, the semi-private score (\lb{}); private scores were not available at the time of writing.

\subsection{Synthetic tasks}

The ARC training set is small, and the highest-scoring systems of 2024 and 2025 trained on generated tasks
\citep{chollet2024arcprize,chollet2026arcprize2025}. re-arc \citep{hodel2024rearc} provides procedural generators for the
ARC-AGI-1 training tasks, which yield unlimited new instances of known rules. BARC \citep{li2025barc} asked an LLM to write
new generator programs from seed programs and used them to train both a program writer (induction) and a model that
outputs grids directly (transduction). NVARC \citep{sorokin2025nvarc} produced synthetic puzzles at a larger scale and
trained its grid model on 3.26M augmented samples. Our generator (\cref{sec:rules}) differs in that each rule is a typed
program whose behaviour on a fixed set of probe grids must differ from that of every other rule, and every task is
replayed by an exact oracle.

\subsection{Grid models and test-time training}

The ARChitects \citep{franzen2024architects} fine-tuned an LLM to output grids as text, trained it again on each test
task's demonstrations at test time, sampled candidates by depth-first search under a probability cut-off, and ranked them
by their likelihood under several symmetry views \citep{franzen2025poe}; this system reached 53.5 points on
the private set of ARC-AGI-1 \citep{chollet2024arcprize}, and their 2025 system replaced the autoregressive model with a masked-diffusion
model \citep{franzen2025architects}. \citet{akyurek2025ttt} showed that test-time training (TTT) on augmented leave-one-out
tasks improves few-shot accuracy on ARC. NVARC follows the ARChitects' design with a Qwen3 \citep{yang2025qwen3} grid model
of at most 4B parameters, trained on synthetic puzzles together with RE-ARC, ConceptARC \citep{moskvichev2023conceptarc}
and MINI-ARC, and per-task LoRA TTT with rank 256. It reported 27.64\% on hidden ARC-AGI-2 tasks in 2025, and the released
model, built on Qwen3-4B-Thinking-2507, later reached 29.72\%. We call this released model
\nvarc{} (\texttt{qwen3\_4b\_grids15\_sft139}). Small recursive networks trained from scratch on ARC data, HRM
\citep{wang2025hrm} and TRM \citep{jolicoeur2025trm}, take a different route; TRM reaches 7.8\% on the ARC-AGI-2 public
evaluation set with 7M parameters.

\subsection{Programs for ARC}

Code-trained language models write programs from natural-language specifications \citep{chen2021codex}.
\citet{greenblatt2024arc} sampled many Python programs per task from GPT-4o and kept those that reproduce the
demonstrations. BARC's induction models write programs directly, and SOAR \citep{pourcel2025soar} fine-tunes the program
writer on its own search traces. A program that reproduces all demonstrations can still be wrong on the test input, so
precision on the test matters as much as the rate of demonstration-perfect programs.

\subsection{Reasoning models and rejection-sampling fine-tuning}

Reasoning models produce a long chain of thought before the answer \citep{guo2025deepseekr1}. Training a model on its own
correct solutions, selected by checking the final answer, is the idea behind STaR \citep{zelikman2022star},
rejection-sampling fine-tuning \citep{yuan2023scaling} and ReST \citep{gulcehre2023rest}. On ARC the check is exact: a trace
is kept only if its final grid equals the known output. \qbig{} \citep{qwen2026qwen38,qwen2026qwen38card} is a dense 27B model with 64
layers, of which 48 use Gated DeltaNet linear attention \citep{yang2025gdn} and 16 use full attention, and a
multi-token-prediction head \citep{gloeckle2024mtp,deepseekai2024v3} that we use for speculative decoding
\citep{leviathan2023speculative,chen2023speculative}. Its chat template exposes a reasoning-effort setting. The ARC Prize
Foundation measured the base model on the semi-private set without the twelve-hour budget: 42.4\% at \texttt{xhigh}, for \$0.45 per
task, and 22.8\% at \texttt{low} \citep{arcprizeleaderboard}.
We serve it with vLLM \citep{kwon2023vllm} and quantize weights to FP8 or to NVFP4, a 4-bit floating-point format with
FP8 scales per block of 16 values \citep{nvidia2025nvfp4,alvarez2025nvfp4blog}, related to the microscaling formats of
\citet{rouhani2023mx}.

\subsection{Mixture-of-experts models and expert pruning}

A sparse mixture-of-experts layer replaces the feed-forward block by many experts and a router that sends each token to a
few of them \citep{shazeer2017moe}; Switch Transformers route each token to one expert and scale the number of experts
at a fixed compute per token \citep{fedus2022switch}, and DeepSeekMoE adds shared experts that every token uses
\citep{dai2024deepseekmoe}. Compute then grows with the active parameters, while memory grows with all of them. Expert
pruning removes experts after training. \citet{lu2024notallexperts} choose the experts to keep by the reconstruction loss of
the layer output on calibration data, and \citet{li2024mcsmoe} merge experts guided by routing statistics. REAP
\citep{lasby2026reap} scores each expert by the gate value times the norm of its output, averaged over the tokens routed
to it, and reports that pruning by this score preserves generative quality better than merging at the same compression.
\qflash{} \citep{qiu2026qwen38next} is a mixture-of-experts reasoning model with Gated DeltaNet layers, 512 routed experts
per layer and 6B of its 125B parameters active per token (\cref{sec:flash-model}).

%% file: sec/rules.tex
\section{A corpus of 12,000 executable rules}
\label{sec:rules}

\subsection{Rules as typed programs}

A rule in our corpus (``law'' in the generator code) is a program in a typed domain-specific language, stored as a JSON
abstract syntax tree and executed by an exact NumPy oracle. The first version of the
language (v1, 2,000 rules) has three stages and one parameter: a \emph{selector} that picks one connected component by a
metric and an objective (25 metrics, minimum or maximum), a \emph{mask} operator applied to that component (38 operators,
48 with their amounts), a \emph{renderer} that draws the result (19 renderers, 27 with their separator widths), and a
connectivity of 4 or 8. This gives $50 \times 48 \times 27 \times 2 = 129{,}600$ distinct programs. The second version (v2,
10,000 rules) has seven families. One, the mask pipeline, extends v1 with one or two post operations on the mask; it spans
88.4M syntactic programs over both connectivities, of which production enumerated connectivity 4 only. The other six have
their own typed stages, for example two selected objects combined by an aligned boolean relation, two framed panels
combined by a 4-bit truth table, the repair of one defect in a symmetric template, rays and floods between role-coloured
walls and sources, and counting a feature over 3 to 8 objects followed by packing tiles along an axis
(\cref{tab:rule-funnel}). \Cref{fig:rule-example} shows one program and the oracle branch that executes it.

Input grids are not written per rule. One scene generator per family proposes inputs for a given program, the oracle
labels them, and gates decide whether the six resulting pairs identify the program. Every input carries a 10-cell
palette key in its border that names the roles of six colours (frame, background, object ink, result, auxiliary,
separator), and a new palette is drawn for every pair, so a rule never depends on absolute colours. The release manifest
records that no language model wrote, selected, labelled or verified any rule; rules are points of the grammar chosen by a
deterministic search. Building the corpus took 55.9 hours from the plan to the generation report on 14 CPU workers, and an
independent audit took a further 26 hours on 8 workers after a first sealing attempt failed; no GPU was used.

\begin{figure*}[t]
\centering
\begin{minipage}{0.95\linewidth}
\begin{lstlisting}[basicstyle=\ttfamily\scriptsize]
# A v2 rule of the mask-pipeline family; its id is the BLAKE2b-128 hash of the canonical JSON.
{"dsl": "unique12000.v2", "family": "mask_pipeline", "family_version": 1,
 "payload": {"connectivity": 4, "selector": {"metric": "vertical_balance", "objective": "max"},
             "base_mask": {"name": "shift_left", "amount": 2},
             "post": [{"name": "branchpoints8"}, {"name": "shift_up", "amount": 1}],
             "renderer": {"name": "panel_h_source_auxiliary", "separator_thickness": 1}}}

def apply_program(grid, program):              # exact oracle, mask-pipeline branch
    ast = parse_program(program)
    scene = parse_scene(grid, ast.connectivity)    # decode, erase palette key
    component = select_component(scene, ast.selector)  # unique winner, ties raise
    mask = apply_mask_op(scene, component, ast.base_mask)
    for op in ast.post:                        # one or two typed post stages
        mask = apply_post_op(mask, scene, component, op)
    return render_program(scene, component, mask, ast.renderer)
\end{lstlisting}
\end{minipage}
\caption{A rule of the mask-pipeline family and the branch of the oracle that executes it (condensed from the generator
source).}
\label{fig:rule-example}
\end{figure*}

\subsection{Admission: behavioural uniqueness and identifiability}

\begin{figure*}[t]
\centering
\footnotesize
\begin{tikzpicture}[
  box/.style={draw, rounded corners=2pt, align=center, text width=#1, inner sep=3pt, minimum height=1.0cm},
  gate/.style={box=#1, fill=blue!6},
  num/.style={box=#1, fill=green!9},
  rej/.style={draw=red!60, rounded corners=2pt, align=left, text width=#1, inner sep=3pt, fill=red!4, font=\scriptsize},
  arr/.style={-{Stealth[length=1.8mm]}, thick},
  node distance=0.42cm]
\node[box=2.45cm] (g) {seven typed v2 grammars (mask pipeline alone: 88.4M programs)};
\node[num=1.9cm, right=of g] (c) {\textbf{89,600}\\ candidates};
\node[gate=3.55cm, right=of c] (b) {behaviour gate: 220 grid probes and 352 typed slots, raw and dihedral vectors new, Hamming $\ge 3$,
$\ge 16$ valid probes};
\node[num=2.2cm, right=of b] (a) {\textbf{37,102}\\ admitted (41.4\%)};
\node[rej=3.55cm, below=0.32cm of b] (r) {rejected: 29,423 dihedral duplicates, 16,957 with fewer than 16 valid probes,
5,705 raw duplicates, 413 near-collisions};
\node[num=1.9cm, anchor=north west] (t0) at ($(g.west |- r.south)+(0,-0.55)$) {\textbf{18,635}\\ rules attempted};
\node[gate=3.3cm, right=of t0] (tg) {task gate for every salt (\cref{alg:task-gate}): replay, materiality, competitors,
shortcuts};
\node[num=2.0cm, right=of tg] (v) {\textbf{13,166}\\ valid (70.7\%)};
\node[box=2.45cm, right=of v] (rel) {release of \textbf{12,000} rules (10,000 v2 + 2,000 v1), audited};
\node[num=2.35cm, right=of rel] (e) {\textbf{95,993} tasks, 575,958 pairs; 600 rules held out};
\draw[arr] (g) -- (c);
\draw[arr] (c) -- (b);
\draw[arr] (b) -- (a);
\draw[arr, red!60, dashed] (b) -- (r);
\draw[arr] (a.south) -- (a.south |- r.south) -- ++(0,-0.25) -| (t0.north);
\draw[arr] (t0) -- (tg);
\draw[arr] (tg) -- (v);
\draw[arr] (v) -- (rel);
\draw[arr] (rel) -- (e);
\end{tikzpicture}
\caption{The admission funnel of the v2 rules (\cref{alg:rules}). Task generation stopped once each family had 1.25 times
its quota of valid tasks, so only 18,635 of the admitted rules were attempted. The release removed aliases, kept each
family's quota and was audited independently; the export added up to seven tasks per rule from new salts.}
\label{fig:rule-pipeline}
\end{figure*}
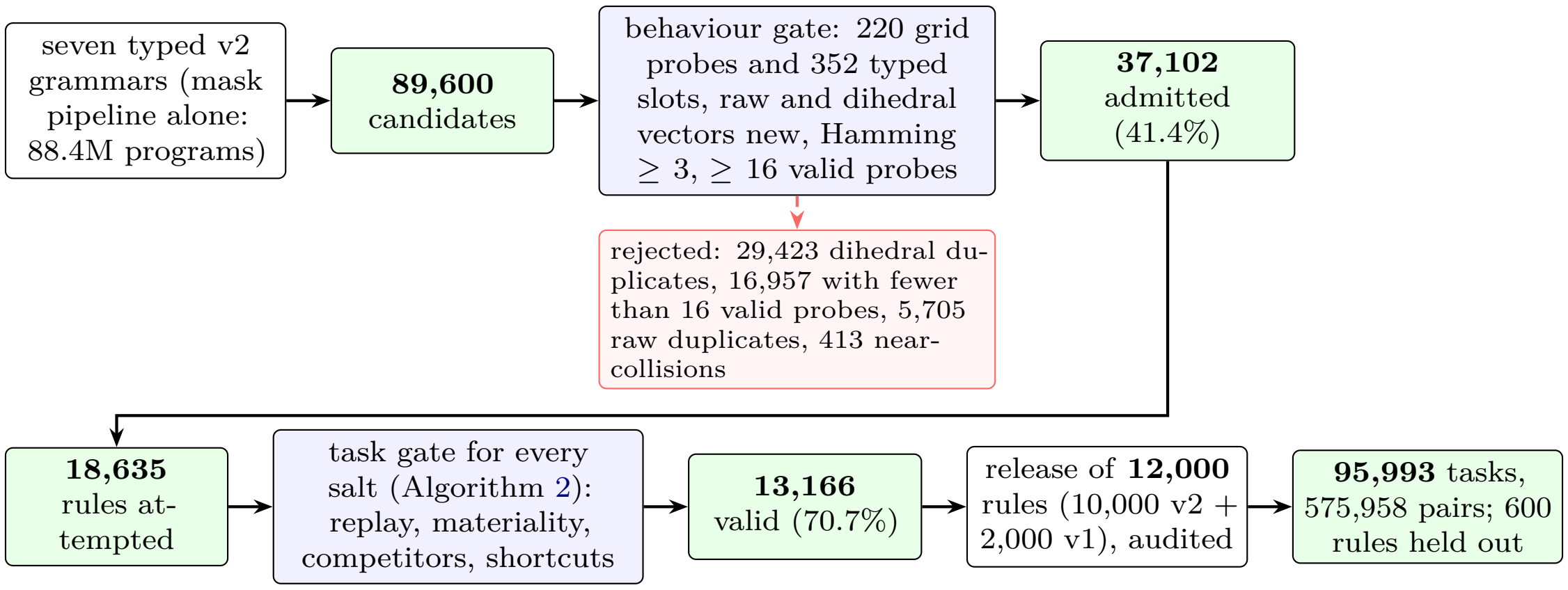

\Cref{alg:rules} gives the construction of the v2 rules and \cref{fig:rule-pipeline} the number of rules that
passed each gate. Two programs can differ syntactically and compute the same
function, so a rule is admitted only if its \emph{behaviour} is new. The behaviour vector has 572 slots: 220 shared probe
grids, 96 panel-boolean slots and 256 template-symmetry slots. Each rule runs only its own segment (the 220 grids, or the
96 or 256 typed inputs of its family) and records in each slot the output or the type of error; the other slots hold a
fixed marker of its family, so rules of different segments always differ. A candidate is admitted if it executes on at
least 16 probes, its vector is new both as is and in its canonical form under the eight dihedral conjugations
$x \mapsto g^{-1}(p(g\,x))$, and its Hamming distance to every admitted rule is at least 3 in both forms. A difference of
signature is a difference of function when two rules give different valid outputs on a shared input; a difference only in
error types or family markers is not, so we call the admitted rules \emph{signature-distinct}. Of 89,600 candidates, 37,102 passed (41.4\%): 29,423 were dihedral-behaviour duplicates, 16,957 had fewer
than 16 valid probes, 5,705 were raw-behaviour duplicates and 413 were near-collisions. No two candidates had the same
syntax tree, so duplicates are a matter of behaviour, not syntax.

A task is accepted for an admitted rule only if its six pairs (five demonstrations and one test) separate the rule from its
nearest competitors. The oracle must replay all six pairs exactly; every typed stage must change the output on at least two
pairs; each of up to 8 shortlisted near rules and every rule that differs in one syntax node must be wrong on at least
two of the six pairs, so that among these competitors any held-out pair is determined by the other five; and a battery of 12
shortcut solvers must solve none of them. The solvers are constant output, identity, shape and colour look-up tables, single
colour swaps, dihedral colour maps, cropping to the non-background region, fixed sub-grids, row selection, column
selection, scaling and tiling, and copying a demonstration output. Task generation stopped once each family had 1.25 times
its quota of valid tasks; of the 18,635 admitted rules that were attempted, 13,166 (70.7\%) produced a valid task. Before
release we removed 199 mask-pipeline rules that reduce to a v1 rule when their post stages are dropped, 41 rules whose
renderer is an exact alias of another, and one rule with a repeated input, and kept the first rules of each family up to
its quota. \Cref{alg:task-gate} gives the task gate as the generator runs it. Its nearest rules are a shortlist
built from hashed projections of the raw and dihedral behaviour vectors and ranked by Hamming distance, not
the result of an exact search over all admitted rules.

The 2,000 v1 rules were built earlier with stricter competitors (all 2,500 programs of the candidate pool) but a
near-collision check only within each selector domain; two pairs of v1 rules lie at dihedral Hamming distance 2 and were
kept as immutable. An independent auditor replayed every gate on the 10,000 v2 rules, treated the v1 rules as fixed, and
found 12,000 distinct identities, 72,000 distinct canonical pairs and no task, pair or input that matches, under dihedral
and colour canonicalisation, any of 19,721 reference tasks (145,053 pairs) from the ARC-AGI data files published by the
ARC Prize Foundation from 2024 to 2026 and three earlier synthetic sets.

\begin{algorithm*}[t]
\caption{Construction of the 10,000 v2 rules and of the 95,993-task export.}
\label{alg:rules}
\begin{algorithmic}[1]
\Require grammars $\mathcal{G}_1, \dots, \mathcal{G}_7$; quotas $q_f$ with $\sum_f q_f = 10{,}000$; candidate multipliers
$m_f$; probe bank $P$ (220 grids, 352 typed slots); the 2,000 v1 rules $L_{v1}$; reference tasks $\mathcal{R}$; seed $\sigma$
\State $C_f \gets$ first $m_f q_f$ programs of a hash permutation of $\mathcal{G}_f$, for each family $f$
\For{each candidate $p$}
  \State $v(p) \gets [\,p(x)\,]_{x \in P}$;\quad $v_8(p) \gets \min_{g \in D_8} [\,g^{-1} p(g\,x)\,]_{x \in P}$
  \State admit $p$ iff $\ge 16$ valid probes, new $v$ and $v_8$, and Hamming distance $\ge 3$ to every admitted rule
\EndFor
\For{admitted rules $p$ and salts $s$, in bounded waves, until each family has $1.25\,q_f$ valid tasks}
  \State propose inputs with the family scene generator (fresh palette per pair, 10-cell palette key); label with $p$
  \State choose 6 pairs with distinct inputs, outputs, target shapes and structures, $\ge 2$ output shapes, $\ge 4$ palettes
  \State accept iff exact replay $\wedge$ every stage material on $\ge 2$ pairs $\wedge$ each of $\le 8$ shortlisted near
  rules and one-node variants wrong on $\ge 2$ pairs $\wedge$ no shortcut solver succeeds
\EndFor
\State drop v1 aliases and renderer aliases; require canonical task, pair and input uniqueness against $L_{v1}$ and
$\mathcal{R}$; keep $q_f$ rules per family
\State independent audit of every gate; seal the release of 12,000 rules
\State per rule, up to 7 further tasks from new salts through the same task gate; at export, replay, duplicate and
leakage checks only, with no new audit or seal (95,993 tasks); hold out 5\% of the rules of each family
\end{algorithmic}
\end{algorithm*}

\begin{algorithm*}[tp]
\caption{Task gate for one admitted rule $p$ (the generator's task worker).}
\label{alg:task-gate}
\begin{algorithmic}[1]
\Require rule $p$ of family $f$; its hard neighbours $N(p)$, at most 8 admitted rules of the same domain (and, for the
mask pipeline, v1 rules), shortlisted by hashed projections of the raw and dihedral behaviour vectors and ranked by
$\min(d_{\mathrm{raw}}, d_{D_8})$ Hamming distance; salts $s_0, \dots, s_1$; scene generator of $f$; the 12 shortcut
solvers $\mathcal{S}$
\State $C \gets \{\, q \in N(p) \cup \mathrm{OneNode}(p) : \mathrm{canon}(q) \ne \mathrm{canon}(p) \,\}$
\Comment{one-node: programs that differ from $p$ in one syntax node}
\For{$s = s_0, \dots, s_1$}
  \State $(x_k, y_k)_{k=1}^{6} \gets$ scene generator$(p, s, N(p))$
  \Comment{fresh palette per pair, distinct inputs, outputs and target shapes, $\ge 2$ output shapes}
  \State \textbf{skip} $s$ if $p(x_k) \ne y_k$ for some $k$ \Comment{exact oracle replay}
  \State \textbf{skip} $s$ if some typed stage of $p$ changes the output on fewer than 2 of the 6 inputs
  \State \textbf{skip} $s$ if $\min_{q \in C} \#\{k : q(x_k) \ne y_k \text{ or } q \text{ fails on } x_k\} < 2$
  \State \textbf{skip} $s$ if a solver in $\mathcal{S}$ solves the test pair from the five demonstrations
  \State \Return the six pairs with their evidence
\EndFor
\State \Return no task for $p$ from these salts
\end{algorithmic}
\end{algorithm*}

\begin{table*}[t]
\centering
\caption{Admission of the 10,000 v2 rules by family. ``Admitted'' rules passed the behavioural identity gate;
``attempted'' rules entered task generation, which stopped at 1.25 times the quota; the yield is relative to the
attempted rules.}
\label{tab:rule-funnel}
\footnotesize
\begin{tabularx}{\linewidth}{@{}l>{\raggedright\arraybackslash}Xrrrrrr@{}}
\toprule
Family & Stages & Programs & Candidates & Admitted & Attempted & Valid tasks & Released \\
\midrule
mask pipeline & select $\to$ mask $\to$ 1--2 post ops $\to$ render & 88.4M & 24,000 & 17,132 & 6,975 & 3,206 (46\%) & 2,400 \\
two objects & two objects $\to$ align $\to$ boolean relation & 104,040 & 14,400 & 6,512 & 3,257 & 2,484 (76\%) & 1,800 \\
multiple objects & group selector $\to$ aggregate & 43,050 & 12,800 & 3,684 & 2,259 & 2,110 (93\%) & 1,600 \\
panel boolean & two panels $\to$ align $\to$ truth table & 50,400 & 9,600 & 4,076 & 1,538 & 1,538 (100\%) & 1,200 \\
template symmetry & repair one defect of a symmetry & 12,960 & 9,600 & 2,311 & 1,709 & 1,500 (88\%) & 1,200 \\
path, ray, flood & rays, shortest paths, floods & 9,360 & 8,000 & 2,102 & 1,613 & 1,307 (81\%) & 1,000 \\
count and pack & rank 3--8 objects $\to$ pack tiles & 49,152 & 11,200 & 1,285 & 1,284 & 1,021 (80\%) & 800 \\
\midrule
total v2 & & & 89,600 & 37,102 & 18,635 & 13,166 (71\%) & 10,000 \\
\bottomrule
\end{tabularx}
\end{table*}

\subsection{Tasks, variants and other generators}

For training we generated up to seven further tasks per rule from new random salts, which gives 95,993 tasks (one rule
yielded only its base task) with 575,958 labelled pairs. Each further v2 task passed the task gate of the base task when it
was generated (exact replay, every stage material on at least two pairs, up to 8 shortlisted near rules and all one-node
variants wrong on at least two pairs, no shortcut solver), and the further v1 tasks passed the v1 gate (materiality, all
programs of the 2,500-program pool wrong on at least two pairs, shortcuts). At export every label was replayed by the
oracle, and duplicate tasks, pairs or inputs shared with the ARC-AGI-2 public evaluation set and leakage across the rule
split were rejected; the export did not rerun the salt search or the materiality, competitor and shortcut gates, and the
95,993-task extension was not independently re-audited or sealed. Four canonical inputs repeat within one split. We held out 5\% of the
rules of each family with all their tasks (600 rules, 4,800 tasks). \Cref{tab:rule-stats} compares the tasks with ARC-AGI-2.
The synthetic inputs are larger (median 750 cells against 361 for the public evaluation set), always contain six colours,
and the output has the shape of the input in 10.2\% of pairs against 69.8\% in the public evaluation set. In the compact
token format of \nvarc{}, the median task takes 5,749 tokens and the longest 11,076.

We also used two other generators. A shape-balanced generator samples 72 short programs so that every query-input size
from $1\times1$ to $30\times30$ appears exactly 30 or 40 times (27,000 and 36,000 tasks). Before the typed corpus, a catalogue of
591 mechanisms in rule-specific Python code, which we call FORGE50, produced a 12,000-task release and a 50,000-task
corpus split by program signature; earlier rule-specific generators also produced 25 families of basic transformations
and 120 declarative program signatures.

\begin{table*}[t]
\centering
\caption{Statistics of the synthetic tasks (95,993-task export, training split) and of ARC-AGI-2. Cells and the share of
pairs whose output has the input's shape are per pair; tokens are per task in the compact format of \nvarc{} (worst
dihedral view).}
\label{tab:rule-stats}
\footnotesize
\begin{tabular}{@{}lrrrrr@{}}
\toprule
Set & Tasks & Input cells & Same shape & Colours/input & Tokens \\
 & & (median) & & (mean) & (median / max) \\
\midrule
v1 rules & 15,200 & 810 & 21.8\% & 6 & 5,994 / 11,076 \\
mask pipeline & 18,240 & 810 & 22.6\% & 6 & 6,175 / 11,076 \\
two objects & 13,673 & 810 & 0\% & 6 & 5,882 / 6,707 \\
multiple objects & 12,160 & 810 & 0\% & 6 & 5,790 / 6,684 \\
panel boolean & 9,120 & 308 & 0\% & 6 & 2,647 / 3,020 \\
template symmetry & 9,120 & 210 & 0\% & 6 & 2,077 / 3,537 \\
path, ray, flood & 7,600 & 644 & 24.3\% & 6 & 6,486 / 9,014 \\
count and pack & 6,080 & 576 & 0\% & 6 & 4,832 / 6,020 \\
\midrule
all synthetic & 91,193 & 750 & 10.2\% & 6 & 5,749 / 11,076 \\
ARC-AGI-2 training & 1,000 & 100 & 65.9\% & 3.90 & 1,130 / 9,330 \\
ARC-AGI-2 public eval. & 120 & 361 & 69.8\% & 5.49 & 3,068 / 9,702 \\
\bottomrule
\end{tabular}
\end{table*}

\subsection{Representative rules and the models trained on them}
\label{sec:rules-examples}

\begin{figure*}[p]
\centering
\includegraphics[width=\linewidth]{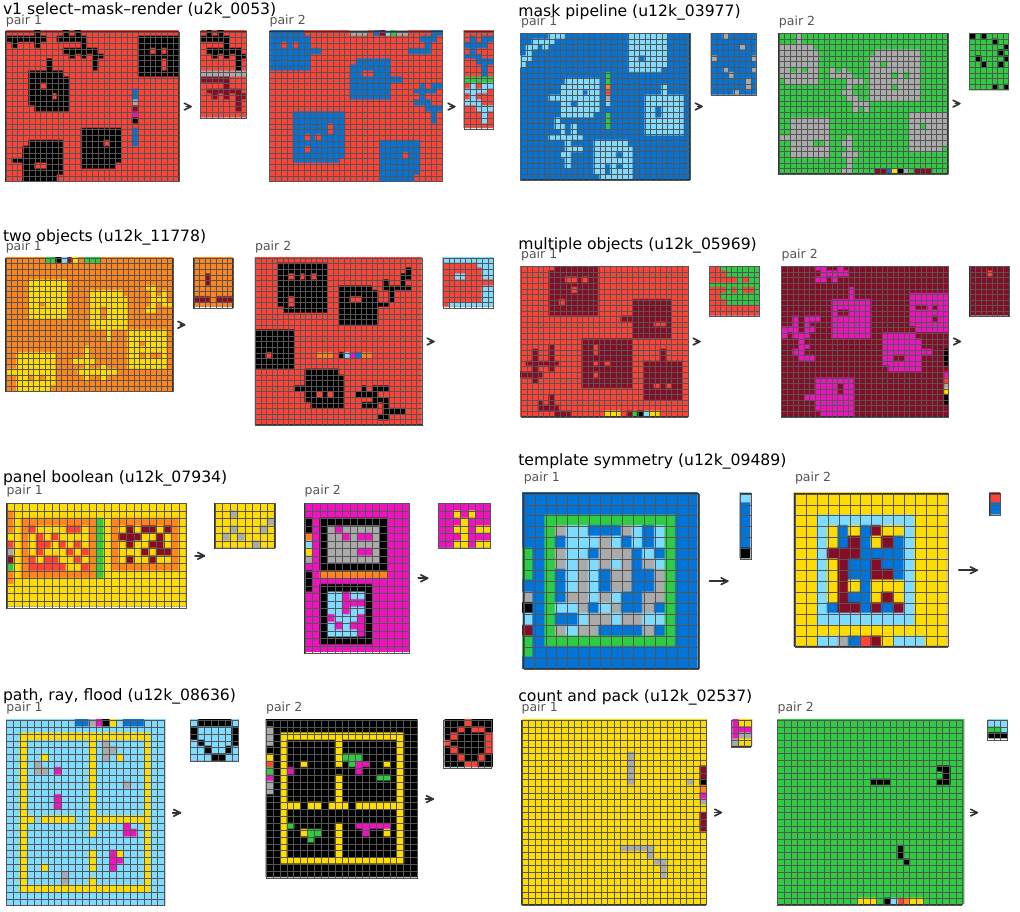}
\caption{The first two of the six pairs of the base task of each corpus rule in \cref{tab:rule-examples}, drawn at one
cell size per panel. Each input carries the 10-cell palette key on one border, and every pair draws a new palette, so the
same rule paints with different colours from pair to pair.}
\label{fig:rule-gallery}
\end{figure*}

\Cref{tab:rule-examples} gives one rule of each corpus family and of the two families of the shape bank, together with the
grid algorithms that its oracle runs, and \cref{fig:rule-gallery} shows two pairs of the base task of each corpus
rule. The descriptions paraphrase the explanations that we computed from the verified
programs (\cref{sec:qwen-synth}); the identifiers name the rules in the release. In every corpus rule, colours are roles
read from the palette key (object, auxiliary, result, background), so the same rule draws with different colours in every
pair. The algorithms are those of the language: connected components in 4- or 8-connectivity with a selector metric,
morphological operations on masks (dilation, halos, hulls, hole filling, endpoints and branch points), alignment and
boolean combination of two or more masks, truth tables over framed panels, symmetry orbits, breadth-first distances for
rays, paths and floods, and counting with binary codes and packing. The shape bank adds whole-grid colour statistics and
orderings of the cells.

Both model lines were trained on these tasks. For \nvarc{}, checkpoints trained on the 1,920 real tasks plus all 95,993
corpus tasks in two views (191,986 synthetic examples) scored 28.61 alone and 27.64 to 32.64 next to the real-task model
(\cref{sec:nvarc-synth,sec:nvarc-ens}); the shape-balanced curriculum of 36,000 tasks was drawn from the same 72 bank programs. For
\qbig{}, 2,231 synthetic tasks from both sources, among them one task for every grid size from $1\times1$ to $30\times30$,
entered the fine-tuning data as explanations or teacher traces (\cref{sec:qwen-synth}).

\begin{table*}[t]
\centering
\caption{One rule of each family. Families above the line are those of the 12,000-rule corpus (v1 and the seven v2
families); the last two are the families of the 72-program shape bank. ``Object'' means a connected component of the
object colour.}
\label{tab:rule-examples}
\footnotesize
\begin{tabularx}{\linewidth}{@{}>{\raggedright\arraybackslash}p{0.17\linewidth}>{\raggedright\arraybackslash}X>{\raggedright\arraybackslash}p{0.25\linewidth}@{}}
\toprule
Family (rule) & Rule & Algorithms \\
\midrule
v1 select--mask--render (u2k\_0053) & Select the object with the smallest anti-diagonal span, shift it one cell up, and draw the
object above its shifted copy with a one-cell separator. & components, selector metric, shift, panel rendering \\
mask pipeline (u12k\_03977) & Select the object with the smallest anti-diagonal span, take the cells that one step of
dilation adds around it, keep those in exactly one of this halo and the object's bounding box, keep the cells with exactly
one occupied neighbour, and crop. & dilation, symmetric difference, endpoint detection \\
two objects (u12k\_11778) & Take the object with the largest vertical imbalance and the topmost object, align their
centres, and keep the cells of the second that the first does not cover. & two selectors, centre alignment, set difference \\
multiple objects (u12k\_05969) & Take the three topmost objects, align their upper-right corners, and keep the cells that
all three cover. & top-$k$ selection, alignment, per-cell consensus \\
panel boolean (u12k\_07934) & Read the two framed panels, flip both top to bottom, align their lower-right corners, and
mark the cells occupied in both. & frame detection, reflection, 4-bit truth table \\
template symmetry (u12k\_09489) & Inside a frame, find the one cell whose change from the object to the auxiliary colour
makes a cross-shaped region symmetric under a top-to-bottom reflection, repair it, and output its symmetry orbit. &
symmetry orbits, unique single-cell repair \\
path, ray, flood (u12k\_08636) & With object cells as walls and auxiliary cells as sources, flood from the source
component that reaches furthest right to distance 2 without crossing walls, and output the halo of the flooded region. &
breadth-first distance, bounded flood, dilation \\
count and pack (u12k\_02537) & Measure the width of every object, order the objects by decreasing perimeter, replace each
by a vertical binary code of its width, and pack the codes from left to right. & counting, ranking, binary encoding,
packing \\
\midrule
spatial logic (bank) & Select the cells of the rarest colour other than the background, replace them by the outline of
their bounding box, add the 180-degree rotation of that outline, and paint it where the input is background. & colour
frequencies, bounding box, rotation, union, occupancy gate \\
shape sequence (bank) & Read the cells from the outside in (first, last, second, second-last, \ldots), append the first
value again if fewer than 30 were read, and output the $(n+1)\times(n+1)$ matrix whose cell $(r,c)$ is foreground exactly
when values $r$ and $c$ differ. & cell ordering, pairwise relation matrix \\
\bottomrule
\end{tabularx}
\end{table*}

\subsection{What is new relative to earlier generators}
\label{sec:rules-new}

\Cref{tab:gen-compare} contrasts the corpus with the generators used by earlier ARC systems. re-arc \citep{hodel2024rearc}
writes one generator per ARC-AGI-1 training task, so it adds new instances of 400 known rules but no new rules. BARC
\citep{li2025barc} and NVARC \citep{sorokin2025nvarc} ask a language model to write new generators or puzzle descriptions,
which adds rules; their descriptions do not report a check that two generators compute different functions or that a
task's pairs determine its rule. Our corpus makes both properties explicit tests, within stated limits. A rule enters only
if its signature on a fixed probe bank differs from that of every admitted rule in at least three of its 572 slots
(outputs or error types), also after canonicalisation under the eight dihedral conjugations; two pairs of v1 rules at
dihedral distance 2 are the exception. The 12,000 rules therefore have 12,000 distinct behaviour signatures, which is
stronger than 12,000 distinct programs but weaker than a proof of 12,000 distinct functions. A task enters only if its six
pairs refute up to 8 shortlisted near rules and every one-node variant of its program on at least two pairs each, every typed stage
changes the output on at least two pairs, and none of 12 shortcut solvers succeeds. Rules are points of a typed grammar
chosen by a deterministic search, every label is replayed by an exact oracle, and no language model wrote, selected or
labelled any rule. Identifiability is relative to this finite set of competitors, not to every possible transformation.
Because the tests are behavioural, they could be applied to other program languages with an executable semantics.

\begin{table*}[t]
\centering
\caption{Synthetic task generators for ARC. ``Distinct rules'' means that rules are checked to behave differently;
``identifiable'' means that each task is checked to separate its rule from a set of competing rules. ``Not reported''
means that we found no such check in the description of the generator.}
\label{tab:gen-compare}
\footnotesize
\begin{tabularx}{\linewidth}{@{}l>{\raggedright\arraybackslash}X>{\raggedright\arraybackslash}Xll@{}}
\toprule
Generator & Source of rules & Checks on rules and tasks & Distinct rules & Identifiable \\
\midrule
re-arc & one hand-written generator per ARC-AGI-1 training task & examples checked by a per-task verifier & by design & not reported \\
BARC & generators written by a language model from seed programs & execution of the generated code & not reported & not reported \\
NVARC & puzzles and code written by a language model & execution of the code & not reported & not reported \\
Ours & points of a typed grammar, deterministic search & signatures on 220 grid probes and 352 typed slots, raw and
dihedral, Hamming distance $\ge 3$ (two v1 pairs at 2); up to 8 shortlisted near rules and one-node variants refuted on
$\ge 2$ pairs; 12 shortcut solvers fail; exact oracle replay & signatures & finite set \\
\bottomrule
\end{tabularx}
\end{table*}

\begin{figure*}[t]
\centering
\includegraphics[width=\linewidth]{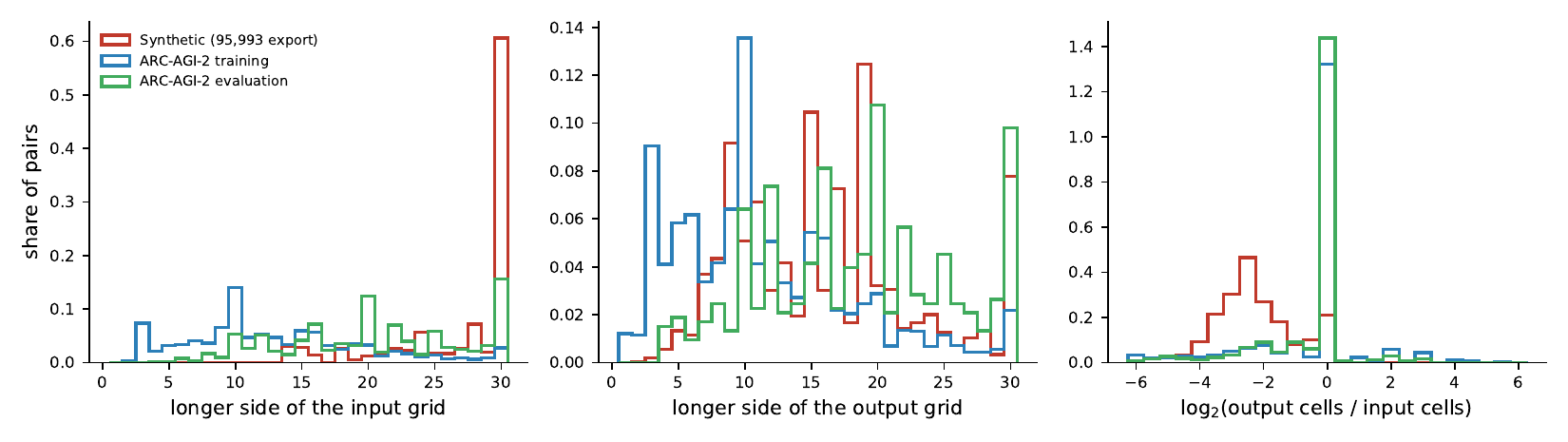}
\caption{Grid sizes of the synthetic corpus (547,158 pairs of the 95,993-task export, training split) and of ARC-AGI-2
(4,308 training and 530 evaluation pairs). Left and middle: the longer side of input and output grids. Right: the ratio of
output to input cells on a $\log_2$ scale; 0 means that the output has as many cells as the input.}
\label{fig:grid-sizes}
\end{figure*}

\paragraph{Grid distribution.} \Cref{fig:grid-sizes} shows the grid sizes of all 547,158 synthetic pairs of the training
split next to ARC-AGI-2. Synthetic inputs are large: the median longer side is 30 cells (quartiles 25 and 30), 89.5\% of
inputs have a side of at least 20 and none has both sides of at most 10, against a median of 20 for the evaluation set
(58.9\% at least 20, 9.4\% at most 10) and 12 for the training set. The outputs are smaller, with a median longer side of 15
cells (quartiles 10 and 19), between the training set (10) and the evaluation set (19). An output has a median of 19\% of the
cells of its input, while in ARC-AGI-2 the median output has the input's size. The scene generators use 105 distinct input
shapes, against 434 in the training set and 169 in the evaluation set.

\paragraph{Difficulty.} The gates make every task hard to solve by a shortcut: the 12 shortcut solvers solve no task by
construction, whereas they solved 15.7\% of the tasks of an earlier rule-specific generator of ours. The tasks are also hard
for the models we trained. Without test-time training, the released \nvarc{} solved 2 of 30 held-out synthetic tasks, and 6
with it; on 18 public evaluation inputs it solved 2 and 3. \qbig{} models fine-tuned for ARC solved 1 of the 30 synthetic
tasks at a 32K thinking cap and 3 to 5 at 64K, while single attempts of the same models at 64K solve 52 to 57 of the 172
public evaluation inputs (30\% to 33\%). A median synthetic task also takes 5,749 tokens in the format of \nvarc{}, against
3,068 for an evaluation task, because its inputs are larger and it has six pairs.

\subsection{How the rules differ from ARC-AGI-2}
\label{sec:rules-compare}

\paragraph{Vocabulary.} A v1 rule reads ``select one object, transform its mask, draw the result''. The 25 selector
metrics measure a connected component: area, bounding-box area, width, height, perimeter, density, number and area of
holes, the four extreme coordinates, the centre row and column, border contacts and touched border sides, the longest row
and column runs, endpoints, branch points, the diagonal and anti-diagonal spans, vertical and horizontal balance, and the
number of symmetries. The 38 mask operators are the identity, bounding-box fills, outlines, complements and corners, object
boundaries and interiors in 4- and 8-connectivity, hole fills, hole masks and hole boundaries, row and column spans, convex
and orthogonal hulls, the seven non-identity dihedral transforms, dilation, erosion and halos, shifts, endpoints, branch
points and the centroid. The 19 renderers extract the result, overlay it on the input, isolate it, place it beside the
source object in a horizontal or vertical panel, or draw it inside a frame. The v2 families add relations between objects
(a boolean relation between two aligned masks, or per-cell support counts over a group of objects chosen by a metric
predicate), truth tables over two framed panels, the repair of one defect against a symmetry or a period, three grid
algorithms (ray casting, the union of all shortest paths, and bounded flooding between walls and sources), and counting a
feature over 3 to 8 objects followed by packing one tile per object along an axis.

\begin{figure*}[tp]
\centering
\includegraphics[width=\linewidth]{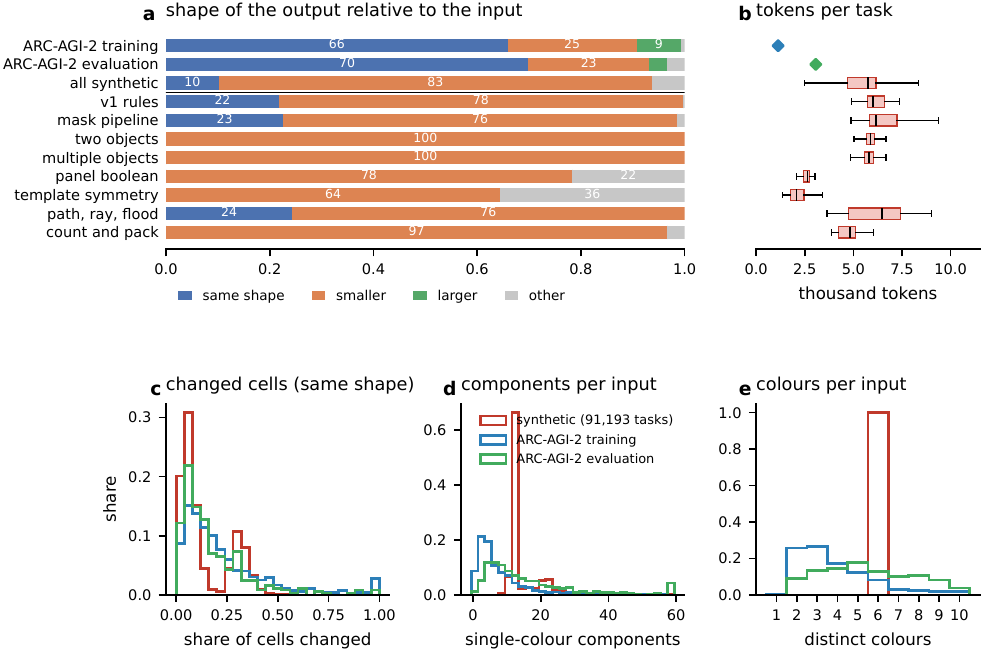}
\caption{Pairs of the training split of the synthetic export (91,193 tasks, 547,158 pairs) against ARC-AGI-2 training
(1,000 tasks) and public evaluation (120 tasks). (a)~Shape of the output relative to the input, by family; numbers are
percentages of pairs. (b)~Tokens per task in the compact format of \nvarc{} (worst dihedral view) for the synthetic rows of
(a), with whiskers at 1.5 times the interquartile range and outliers not drawn; diamonds are the ARC-AGI-2 medians of
\cref{tab:rule-stats}. (c)~Share of changed cells in pairs whose output has the input's shape. (d)~Single-colour
4-connected components per input outside its most frequent colour, palette key included, clipped at 59. (e)~Distinct
colours per input.}
\label{fig:rule-dists}
\end{figure*}

\paragraph{Grids and pairs.} \Cref{tab:rule-vs-arc} compares the pairs of the training split with ARC-AGI-2, and
\cref{fig:rule-dists} shows their distributions. Three
differences follow from the design. First, no synthetic output is larger than its input and 83.5\% are smaller, because
the renderers draw a selected object, a panel or a packed strip; the median output has 19\% of the input's cells, whereas
the median ARC-AGI-2 output has the input's size, and 8.6\% of training and 3.4\% of evaluation outputs are larger, most of
the training ones by integer factors. Cropping is also rare: 0.1\% of the smaller synthetic outputs are a sub-grid of the
input, against 23.7\% in the training set, because the shortcut battery rejects tasks that cropping solves. Second, the
v2 pair selection requires at least two output shapes, so only 0.2\% of synthetic tasks (173, all v1) keep one output
shape across their pairs,
against 41.3\% of training and 20.0\% of evaluation tasks. Third, colours carry no fixed meaning. A fresh palette is drawn
for every pair, so no synthetic task keeps the same most frequent colour across its inputs, against 74.5\% of training and
46.7\% of evaluation tasks, and black is the most frequent colour of 10.0\% of synthetic inputs against 66.8\% and 34.5\%.
Synthetic outputs never contain a colour absent from their input, which 30.2\% of training and 12.1\% of evaluation outputs
do. Where the output keeps the input's shape, synthetic rules change fewer cells (median 7.9\% against 16.0\% and 12.3\%).
The inputs are about as crowded as evaluation inputs, with a median of 12 single-colour components outside the most
frequent colour, part of them from the palette key, against 6 in training and 12 in evaluation inputs.

\begin{table*}[t]
\centering
\caption{Transformation statistics of the synthetic tasks (95,993-task export, training split) and of ARC-AGI-2. Shares
are of pairs unless the row says tasks; a sub-grid is a contiguous block of the input. Components are 4-connected
single-colour regions of all colours except the input's most frequent one; the synthetic count includes the cells of the
palette key.}
\label{tab:rule-vs-arc}
\footnotesize
\begin{tabularx}{\linewidth}{@{}>{\raggedright\arraybackslash}Xrrr@{}}
\toprule
Statistic & Synthetic & ARC-AGI-2 training & ARC-AGI-2 eval. \\
 & (91,193 tasks) & (1,000 tasks) & (120 tasks) \\
\midrule
Pairs per task (mean) & 6 & 4.31 & 4.42 \\
Output has the input's shape & 10.2\% & 65.9\% & 69.8\% \\
Output smaller (neither side longer) & 83.5\% & 24.9\% & 23.4\% \\
Output larger (neither side shorter) & 0.0\% & 8.6\% & 3.4\% \\
Other shape changes & 6.3\% & 0.7\% & 3.4\% \\
Output cells over input cells (median) & 0.19 & 1.00 & 1.00 \\
Smaller outputs that are a sub-grid of the input & 0.1\% & 23.7\% & 0.8\% \\
Tasks with one output shape for all pairs & 0.2\% & 41.3\% & 20.0\% \\
Changed cells in same-shape pairs (median) & 7.9\% & 16.0\% & 12.3\% \\
Outputs using only colours of their input & 100\% & 69.8\% & 87.9\% \\
Colours per input (mean) & 6.00 & 3.90 & 5.49 \\
Tasks whose inputs share the most frequent colour & 0\% & 74.5\% & 46.7\% \\
Inputs whose most frequent colour is black & 10.0\% & 66.8\% & 34.5\% \\
Components per input (median) & 12 & 6 & 12 \\
\bottomrule
\end{tabularx}
\end{table*}

\paragraph{The design rules of ARC-AGI-2.} \citet{chollet2025arcagi2} describe four properties that ARC-AGI-2 tasks were built
to require: several rules that act at once and interact, several steps where each depends on the previous one, rules whose
application depends on a cue in the grid, and symbols whose meaning is defined inside the task. Our rules cover the first
two only in a fixed form. A rule composes typed stages in one order (select, transform, draw, or the stages of its
family), and a task applies exactly one rule. Stages run in a fixed order with no conditional or recursive composition;
some stages iterate inside a fixed algorithm (paths, rays and floods, hole filling, component labelling), and the
input-dependent choices are the object that a selector picks, the source of a path, the ranking of a count pack and the
unique repair that the template family finds by search. The palette key is a symbol definition, but always of the same six roles, in the same 10-cell strip at the
border in one of four orientations; a model can learn to read it once, while a legend in an ARC-AGI-2 task has a new form
in every task. Two ingredients of real tasks have no counterpart in the vocabulary: colours with a meaning of their own,
because every rule refers to colour roles, and outputs larger than the input, as in tiling, scaling or drawing beyond an
object.

\subsection{Limitations of the corpus}
\label{sec:rules-limits}

The gates guarantee that each task separates its rule from a finite shortlist of near rules of the corpus and from the
one-node variants of its program, not that a person, or a model
trained on real ARC tasks, can infer it; we ran no human or model inferability test. The format differs from ARC-AGI-2 in
ways that a model can learn as shortcuts: the palette key appears in every input, every input has six colours, and most
outputs are smaller than their input (\cref{sec:rules-compare}). Only 4,400 of the 12,000 rules (the v1 rules and the mask-pipeline family) have a rule-specific
English description; the other 7,600 carry a placeholder. Of the 600 held-out rules, 457 reappear in one derived training
set and, by our notes, 498 in a standalone program set, although no pair does.
\Cref{sec:nvarc,sec:qwen} report the consequences: the grid model fitted synthetic tasks when trained on them,
but no model trained with them was better on real tasks, alone or next to the real-task model, and the best \nvarc{}
configuration uses no checkpoint trained on the corpus (\cref{sec:nvarc-ens}).

%% file: sec/nvarc.tex
\section{Real and synthetic training data for the NVARC grid model}
\label{sec:nvarc}

\subsection{The base pipeline}
\label{sec:nvarc-base}

\nvarc{} is a Qwen3-4B model with its vocabulary cut to 16 tokens, which include the ten colours, a newline, role
markers, padding and end of sequence; it was fully fine-tuned by its authors on 3.26M samples from 104,539 puzzles,
about 90\% of them synthetic \citep{sorokin2025nvarc}. The released pipeline solves each task in five steps, which we kept
(\cref{alg:nvarc}). First, a fresh LoRA adapter (rank 256, $\alpha=32$, rank-stabilised, on the seven projection types
plus the embedding and output layers) is trained for one epoch on 128 augmented copies of the task's demonstrations (8
dihedral transforms $\times$ 16 colour permutations, demonstrations shuffled), with learning rate $5\times10^{-5}$.
Second, every test input is rendered in 16 views (8 dihedral transforms $\times$ 2 colour permutations). Third, a
depth-first search over the 12 legal tokens keeps every branch whose cumulative probability stays above 0.2. Fourth, each
distinct candidate, mapped back to the original frame, is scored by its negative log-likelihood (NLL) under 8 fresh views.
Fifth, the score of a candidate is its number of search hits minus its mean rescoring NLL, and the top two are the two attempts.
On four L4 GPUs, one worker per GPU, test-time training takes 193 to 870 seconds per task and a whole task 517 to 1,371
seconds. Four runs with the base model, three of the released pipeline unchanged and one in our runtime with a
12,288-token context, scored 28.47, 29.86, 29.03 and 28.47 on the semi-private set (mean 28.96, standard deviation
0.66); the three unchanged runs alone average 29.12. For every run of this section we read the checkpoints from the
source code or the log of the version that produced the score; a fifth run that we had first counted as a base run loads
a checkpoint of our own fine-tuning and is counted in \cref{sec:nvarc-real}.

\begin{algorithm*}[t]
\caption{Fine-tuning and test-time pipeline for \nvarc{}. Lines 1--4 run once offline; lines 5--12 run per task on each
of the four GPUs.}
\label{alg:nvarc}
\begin{algorithmic}[1]
\Require base model $M_0$; offline corpus $D$; hidden test tasks $\mathcal{T}$; deadline 12 h
\State $D \gets$ concatenation of ARC-AGI-1 train+eval, ARC-AGI-2 train and ARC-AGI-2 public eval (1,920 records; test
pairs appended as demonstrations)
\State for each task, 16 views: 8 dihedral transforms $\times$ 2 colour bijections, demonstrations shuffled
\State train rank-256 LoRA (+ embedding and output layers), cross-entropy on every output grid, learning rate
$5\times10^{-5}$ cosine, batch 1, 8,192 tokens $\Rightarrow M_1$
\State optionally continue $M_1$ on real plus synthetic tasks $\Rightarrow M_2$; place a checkpoint on each GPU
\For{each task $\tau$ popped from a shared queue ordered by estimated cost}
  \State reset the LoRA adapter; build 128 augmented sequences of $\tau$'s demonstrations; train one epoch
  \For{each test input $x$ of $\tau$}
    \State render 16 views of $x$; depth-first search keeping branches with cumulative probability $\ge 0.2$
    \State for each distinct de-augmented grid $y$: $s(y) \gets$ mean NLL of $y$ under 8 fresh views
    \State score$(y) \gets$ hits$(y) - s(y)$; attempts $\gets$ two highest scores
  \EndFor
\EndFor
\State write the answer file; unanswered inputs get \texttt{[[0]]}
\end{algorithmic}
\end{algorithm*}

\subsection{Continued fine-tuning on real tasks}
\label{sec:nvarc-real}

Before any per-task training, we continued to fine-tune \nvarc{} on all real tasks we could label: the 800 ARC-AGI-1
training and evaluation tasks, the 1,000 ARC-AGI-2 training tasks and the 120 ARC-AGI-2 public evaluation tasks with
their test outputs appended as further demonstrations. This gives 1,920 source records with 6,427 demonstration pairs
(3,852 distinct), rendered in 16 views per record (30,720 examples). The records cover 1,147 task identifiers: 767
ARC-AGI-1 tasks reappear among the ARC-AGI-2 training tasks, mostly with the same demonstrations in another order, and 6
among the evaluation tasks with one more demonstration, so these tasks carry twice the weight of the others. We trained a rank-256 LoRA adapter with the same targets as test-time training, learning rate
$5\times10^{-5}$ with a cosine schedule over two planned epochs (61,440 steps), batch 1 and an 8,192-token context, on one
L4. The run stopped after 10 hours at step 34,273 (1.11 epochs); the checkpoint at step 34,000 is the model we call
\emph{ckpt-34000}. Training on the public evaluation tasks makes local evaluation on them meaningless for this model and
all its descendants, but it does not touch the hidden semi-private tasks.

Twenty-two hidden runs used ckpt-34000, or a checkpoint of the same training run saved 2,000 or 4,000 steps earlier, as
the only model on all four GPUs, and averaged 31.72 on the semi-private set (standard deviation 1.52,
\cref{tab:nvarc-lb}). The one run of ckpt-34000 in the otherwise unchanged released pipeline scored 31.53, and the best
of the 22 scored 34.44. A half-epoch checkpoint of an earlier one-epoch run on the same records scored 29.03.

\subsection{Synthetic tasks}
\label{sec:nvarc-synth}

We added synthetic tasks from several sources. Continuing ckpt-34000 with 25\% of its episodes drawn from the FORGE50
generator (a catalogue of rule-specific Python generators, separate from the typed corpus of \cref{sec:rules}; learning
rate $10^{-5}$) gave a model that scored 29.31 alone, a single draw 2.4 points below the ckpt-34000 mean. A second FORGE50 recipe, a weighted mixture of 86\% FORGE50 episodes, 12\% real
ARC tasks from three pools including the public evaluation set (5\%) and 2\% hand-made tasks, scored 28.47. A curriculum of 36,000 tasks from the shape-balanced generator
plus 1,090 real tasks, started from an adapter trained on a 6,000-task mix of real and rule-corpus tasks, ran for 47.8
hours on one RTX~3090~Ti. It raised accuracy on new tasks from the 72 programs of its own generator from 5.8\% to 61.7\% (14
to 148 of 240) and changed accuracy on 30 real evaluation tasks from 9 to 8 of 47 inputs; on the semi-private set it scored 27.22 at
step 5,500 and 26.53 at the end. Training from the base model on 1,920 real tasks, 3,600 shape-balanced tasks and 1,980
rule-corpus tasks for 120,000 steps scored 28.06. We also fine-tuned checkpoints from the base model on the real tasks
together with all 95,993 tasks of the rule corpus in two views each (191,986 synthetic examples; rank 256, one epoch, on an
RTX~5090). One of them, run alone on all four GPUs, scored 28.61, 3.1 points below the ckpt-34000 mean; next to
ckpt-34000 on other GPUs, these checkpoints scored 27.64 to 32.64 in five runs (mean 30.53, \cref{sec:nvarc-ens}).

About 25 shorter controlled runs on an RTX~5090 or an RTX~3090~Ti tested data mixtures and losses: 1,000 to 97,113 tasks
from the rule corpus and real data, plain cross-entropy, a likelihood-threshold ``gate'' loss matched to the 0.2 search
cut-off, a hard-tail weighting, a KL anchor to the base model's correct tokens, colour-consistency regularisation, and one
to eight views per task. On 15 public evaluation tasks (18 inputs, 16 views, greedy decoding, no test-time training),
they solved 1 to 3 inputs against 2 for the base model. In the same setting, per-task test-time training alone solved 3,
and on 30 held-out synthetic tasks it solved 6 against 2 (bootstrap 95\% interval for the difference $+3.3$ to $+26.7$
points).

\subsection{Runtime changes and checkpoint ensembles}
\label{sec:nvarc-ens}

Our best \nvarc{} configuration (best draw 34.86) keeps the algorithm of \cref{alg:nvarc} and the model ckpt-34000, and
changes how the four GPUs and the twelve hours are used. We recovered the executed source of this configuration together
with its log and per-GPU task statistics:
\begin{itemize}
\item every GPU loads ckpt-34000, with a context length of 13,312, 12,288, 10,240 and 8,192 tokens on GPUs 0 to 3 instead
of 8,192 on all four, so the GPUs differ only in how many demonstrations fit into the prompt;
\item tasks are taken from one shared queue in order of estimated cost, cheapest first, and each task gets a time limit
between 1,200 and 3,000 seconds computed from the remaining budget and queue, instead of a fixed 1,200 seconds;
\item the global deadline is 12 hours plus 10 minutes instead of 12 hours minus 10 minutes;
\item search scores are kept in FP32 instead of BF16, and the rescoring views use a deterministic seed per task;
\item a worker that fails on three consecutive tasks exits and the others take its tasks.
\end{itemize}
Each worker solves whole tasks, and there is no vote across GPUs. This configuration therefore contains no checkpoint
trained on our synthetic data; its gain over the base model comes from continued fine-tuning on real tasks and from the
runtime changes.

Five runs of this configuration scored 33.61, 34.86, 33.06, 33.19 and 30.56 (mean 33.06, standard deviation 1.57),
4.10 points above the base model (Welch $t=5.3$) and 1.33 points above the mean of ckpt-34000 in earlier runtimes
($t=1.7$), a difference within the run-to-run variation. In the same code, a configuration that ran the FORGE50
continuation in place of ckpt-34000 on the fourth GPU (with a 13,312-token context) scored 34.31, 33.06 and 30.56 (mean 32.64), 0.41 points lower. In
36 runs whose checkpoint lists we read from the source or log of each version, ensembles of ckpt-34000 with other checkpoints averaged
30.62; the five of them that included a checkpoint trained on the real tasks plus the whole rule corpus averaged 30.53.
No model trained with synthetic tasks that we ran alone scored above the real-task model (29.31, 28.61, 28.47, 28.06,
27.22 and 26.53 against a mean of 31.72), and placing a synthetic-trained checkpoint next to ckpt-34000 did not raise the
mean score.

\begin{table*}[t]
\centering
\caption{Semi-private scores of the \nvarc{} line (ARC-AGI-2, 120 semi-private tasks, two attempts). Draws are separate
runs; the pooled standard deviation of repeated runs of identical configurations is 1.94 points (27 draws of 9
configurations). Checkpoint lists come from the source code or log of the version that produced each score. $\Delta$
is the difference of means to the base model. Rows marked $\dagger$ ran in the same
runtime as the row of ckpt-34000 with runtime changes, against which their differences are $-5.8$ to $-6.5$ (shape-balanced) and $-5.0$ (real and
synthetic).}
\label{tab:nvarc-lb}
\footnotesize
\begin{tabularx}{\linewidth}{@{}>{\raggedright\arraybackslash}Xrrrr@{}}
\toprule
System & Draws & Mean $\pm$ SD & Best & $\Delta$ \\
\midrule
Base model (three runs of the released pipeline, one in our runtime) & 4 & 28.96 $\pm$ 0.66 & 29.86 & -- \\
\quad frozen base sampled without test-time training or search, \qbig{} selects & 1 & 10.42 & 10.42 & $-18.54$ \\
ckpt-34000 or a checkpoint of the same run, continued on 1,920 real tasks & 22 & 31.72 $\pm$ 1.52 & 34.44 & $+2.77$ \\
\quad + 25\% FORGE50 synthetic episodes (one model) & 1 & 29.31 & 29.31 & $+0.35$ \\
FORGE50 mixture with 86\% synthetic episodes (one model) & 1 & 28.47 & 28.47 & $-0.49$ \\
Base + 1,920 real + 95,993 rule-corpus tasks (one model) & 1 & 28.61 & 28.61 & $-0.35$ \\
Base + 1,920 real + 5,580 synthetic tasks$^\dagger$ & 1 & 28.06 & 28.06 & $-0.90$ \\
Shape-balanced curriculum (36,000 tasks)$^\dagger$ & 2 & 26.88 $\pm$ 0.49 & 27.22 & $-2.08$ \\
Ensembles of ckpt-34000 and other checkpoints & 36 & 30.62 $\pm$ 2.08 & 34.31 & $+1.66$ \\
\quad with a checkpoint trained on 1,920 real + 95,993 rule-corpus tasks & 5 & 30.53 $\pm$ 1.93 & 32.64 & $+1.57$ \\
ckpt-34000 on all four GPUs, + runtime changes & 5 & \textbf{33.06 $\pm$ 1.57} & \textbf{34.86} & $\mathbf{+4.10}$ \\
\quad FORGE50 continuation on the fourth GPU, same code & 3 & 32.64 $\pm$ 1.91 & 34.31 & $+3.69$ \\
\bottomrule
\end{tabularx}
\end{table*}

\begin{table*}[t]
\centering
\caption{Offline fine-tuning runs of \nvarc{}. Real tasks come from ARC-AGI-1 and ARC-AGI-2; synthetic tasks come from
FORGE50, the shape-balanced generator and the 12,000-rule corpus of \cref{sec:rules}. All runs use rank-stabilised LoRA.
Steps are optimizer steps of one task (rank 256) or of four tasks (rank 64).}
\label{tab:nvarc-runs}
\footnotesize
\begin{tabularx}{\linewidth}{@{}llr>{\raggedright\arraybackslash}Xrrrl@{}}
\toprule
Model & Init & Real & Synthetic & Views & Rank & Steps & Hardware, time \\
\midrule
ckpt-34000 & base & 1,920 & -- & 16 & 256 & 34,000 & 1 $\times$ L4, 10.0 h \\
FORGE50 continuation & ckpt-34000 & 1,920 & FORGE50, 25\% of episodes & 16 & 256 & 45,000 & rented GPU \\
mixed6000 & base & 1,090 & 4,910 rule corpus & 2 & 64 & 1,500 & RTX 3090 Ti \\
shape37090 & mixed6000 & 1,090 & 36,000 shape-balanced & 2 & 64 & 9,273 & RTX 3090 Ti, 47.8 h \\
original7500 & base & 1,920 & 3,600 shape + 1,980 rule & 16 & 256 & 120,000 & RTX 5090 \\
all-train & base & 1,120 & 95,993 rule corpus & 1 & 256 & 12,140 & RTX 5090 \\
real + corpus & base & 1,920 & 95,993 rule corpus & 2 & 256 & 1 epoch & RTX 5090 \\
\bottomrule
\end{tabularx}
\end{table*}

\paragraph{What did not work.} Sampling the frozen base model without test-time training or search, with \qbig{} choosing
two of the candidates, scored 10.42. Within test-time training, a learning rate of $10^{-5}$ instead of $5\times10^{-5}$
scored 25.69, 160 sequences for two epochs with a 0.19 search cut-off scored 20.00, and a reduced profile (64 training
sequences, 8 inference views, 4 rescoring views, 720 seconds per task) scored 12.50 in a run that left part of the inputs
unanswered. Merging the adapter of ckpt-34000 into BF16 weights and running it at longer contexts scored 28.61 and 28.19,
against 31.1 to 31.5 for the adapter loaded at 8,192 tokens, although merge, context and code changed together. A short
4-task check run, which we used to check every configuration before a hidden run, did not predict the semi-private score:
over 49 runs the Pearson correlation between its score and the \lb{} score was 0.14.

%% file: sec/qwen.tex
\section{Fine-tuning \qbig{} on its own answer-verified reasoning traces}
\label{sec:qwen}

This section describes the main line of work: rejection-sampling fine-tuning (RFT) of the dense \qbig{} reasoning model
\citep{qwen2026qwen38card} on traces that it generated itself and whose answers match the known outputs, and the inference
procedure that we built around the fine-tuned model. The line took the semi-private score of \qbig{} from 26.81 to 44.03,
the level that the ARC Prize Foundation measured for the base model without a time limit (42.36). We first describe
two fixes that the base model needed (\cref{sec:qwen-fixes}), then the trace pipeline (\cref{sec:qwen-traces}), the training
recipe (\cref{sec:qwen-train}), the later data variants (\cref{sec:qwen-variants}) and the inference procedure
(\cref{sec:qwen-infer}). \Cref{sec:qwen-eval} compares the models on the public evaluation set without a time limit
and, where they were run there, on the semi-private set.

\paragraph{Evaluation sets.} Unless noted, offline numbers in this section are on the ARC-AGI-2 public evaluation set, 120
tasks with 172 test inputs. We report the number of exactly correct test inputs and, where noted, the task score (the mean
over tasks of the fraction of solved inputs, times 100). Because the RFT training data include traces for some evaluation
tasks, we also use three subsets: \emph{H}, 30 randomly chosen tasks (45 inputs) that no model of the RFT line except FINAL
was trained on; \emph{T}, 61 tasks (90 inputs) whose correct base-model traces are in the \rftv{1} data; and \emph{N}, 29
tasks (37 inputs) that no model had solved at the time the subset was defined. H is the only random subset that is
disjoint from the training data of the RFT line before FINAL; we used it repeatedly during development, so it is a
development subset rather than an untouched test set. Unless stated otherwise,
decoding uses $T=0.6$, top-$p=0.95$, top-$k=20$ and the \texttt{xhigh} reasoning effort of the chat template.

\subsection{Two fixes to the base model}
\label{sec:qwen-fixes}

\paragraph{End-of-turn token inside the thinking block.} \qbig{} samples the end-of-turn token (id 248046) in the middle of
its reasoning, typically while it enumerates grid rows. The serving engine then returns an empty answer. At low effort on a
ten-task positive-control set (\emph{parity10}, 13 inputs that the evaluation of the ARC Prize Foundation reports as solved), 11 of 26
requests ended this way, and on two tasks every attempt did. At the stopping step the token had probability between 5\%
and 62\%. The failure persisted with a BF16 output head, a BF16 key-value cache, a different attention backend and two
other checkpoints; all of these runs used quantized weights, so we did not test unquantized ones, but nothing points to the
quantization. We split each attempt into two requests on token ids: a thinking request in which the end-of-turn and
end-of-text tokens carry a logit bias of $-100$ and which stops at \texttt{</think>}, followed by an answer request under a
regular-expression grammar for grids. This raised task-averaged two-attempt accuracy on parity10 from 36.7\% to 66.7\%
(5 to 8 of 13 inputs), and valid grids from 15 to 26 of 26. Because the serving engine rejects logit biases when speculative
decoding is enabled, the production checkpoints instead have the two rows of the output head set to zero; every
fine-tuned export below carries this patch.

\begin{figure*}[tp]
\centering
\includegraphics[width=\linewidth]{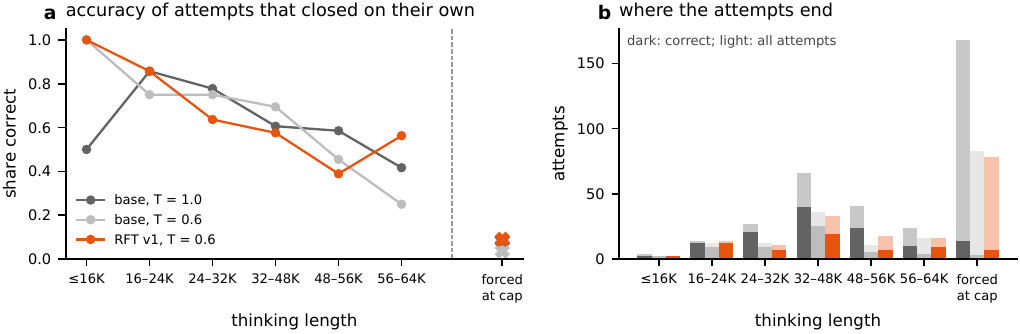}
\caption{Single attempts of \qbig{} on the public evaluation set at a 64K cap (the runs of \cref{tab:qwen-length}).
(a)~Share correct among natural completions, by thinking length, and among forced completions at the cap
(crosses). (b)~Attempts in each bucket (light) and correct attempts (dark).}
\label{fig:qwen-length}
\end{figure*}

\paragraph{Effort and forced completions.} With the fix in place, the reasoning effort decides most of the accuracy. On the
semi-private set, low effort scored 9.86 and \texttt{xhigh} with a 64K-token thinking cap scored 26.81 (twice, in two
draws of the same configuration). Offline, \texttt{xhigh} with two 64K attempts at $T=1.0$ solved 84 of 172 inputs (task score
49.86), but 168 of the 344 attempts reached the cap. We call an attempt a \emph{natural completion} when the model ends its
thinking before the cap, and a \emph{forced completion} when the engine closes the thinking at the token cap. Forced
completions are rarely right: 14 of 168 (8\%), against 42\% to 86\% for natural completions in each length bucket up to 64K
(\cref{tab:qwen-length,fig:qwen-length}). At low effort, accuracy collapses with length, from 72\% for attempts below 16K tokens to 3\%
above 48K; at \texttt{xhigh} it declines more slowly, and length is confounded with task difficulty. Under the twelve-hour
limit the problem is therefore how to spend tokens: long caps leave inputs unanswered (uniform 88K caps scored 18.89), and
short caps turn natural completions into forced ones.

\begin{table*}[t]
\centering
\caption{Accuracy of single attempts by thinking length on the public evaluation set (172 inputs per run, \texttt{xhigh},
64K cap, 2-request harness: thinking, then a separate answer request). Columns are thinking lengths of natural completions;
the last column counts forced completions at the cap. Entries are correct/attempts.}
\label{tab:qwen-length}
\footnotesize
\begin{tabular}{@{}lccccccc@{}}
\toprule
Run & $\le$16K & 16--24K & 24--32K & 32--48K & 48--56K & 56--64K & forced at cap \\
\midrule
Base, $T=1.0$, 344 attempts & 2/4 & 12/14 & 21/27 & 40/66 & 24/41 & 10/24 & 14/168 \\
Base, $T=0.6$, 172 attempts & 2/2 & 9/12 & 9/12 & 25/36 & 5/11 & 4/16 & 3/83 \\
\rftv{1}, $T=0.6$, 172 attempts & 2/2 & 12/14 & 7/11 & 19/33 & 7/18 & 9/16 & 7/78 \\
\bottomrule
\end{tabular}
\end{table*}

\subsection{Self-generated answer-verified traces}
\label{sec:qwen-traces}

\paragraph{Answer-only training does not transfer.} Our first fine-tune taught the model to emit the answer grid with
thinking disabled: 37,201 examples (33,915 shape-balanced synthetic tasks, 2,085 tasks from the 95,993-task corpus of
\cref{sec:rules}, 1,076 official training inputs and 125 public evaluation inputs), LoRA rank 32, learning rate
$2\times10^{-5}$, 3,149 steps in about 21 hours. It scored 2.08 on the semi-private set, close to the thinking-off score that the
ARC Prize Foundation reports for the base model \citep[1.5\% on the same set;][]{arcprizeleaderboard}. A continuation on all 1,147
labelled official tasks in 16 symmetry and colour views solved 0 of 9 inputs of five tasks that were in its own training
data. We concluded that the targets must keep the reasoning and that the prompt during training must be the inference
prompt, token for token.

\paragraph{Rejection-sampled traces (\rftv{1}).} We then sampled the base model on every labelled training input. The
generator was \qbig{}-FP8 with the end-of-turn patch, three-token multi-token prediction (MTP) and an FP8 key-value cache,
served by vLLM \citep{kwon2023vllm} on one RTX~PRO~6000 (96~GB). The prompt was the prompt of the ARC Prize Foundation's
benchmarking harness, with JSON grids, rendered by the chat template at \texttt{xhigh}. Each input received two attempts (one in a 100-task pilot) with a
48,000-token thinking cap; at the cap the sentence ``Considering the limited time by the user, I have to give the solution
based on the thinking directly now.'' and \texttt{</think>} were appended, and the answer was decoded under the grid
grammar. \Cref{alg:rft} gives the procedure. The base model solved 959 of the 1,076 ARC-AGI-2 training inputs (89.1\%;
1,000 tasks). To these we added 170 correct traces for public evaluation tasks from earlier base-model runs at 64K and 88K
caps ($T=0.9$ to $1.0$), excluding the 30 H tasks.

A trace was kept only if its grid equals the label, it is a natural completion, it contains no chat tags and no
give-up phrase (``best guess'', ``I'm stuck'') in its last 6,000 characters, it is at most 220K characters long, no line of
30 or more characters repeats six times or more, at least 60\% of its lines are unique, and its zlib compression ratio is at
least 0.18 (a loop detector). Per input we kept the shortest surviving trace for training tasks and the two shortest for
evaluation tasks. The result, \texttt{sft\_main1}, has 1,036 rows (900 from 846 training tasks and 136 from 61 evaluation
tasks) and 17.08M thinking tokens; the median trace has 12,159 thinking tokens, the 90th percentile 37,962, and the
evaluation rows have a median of 38,481. A correct trace of at most 16K tokens exists for 619 of 935 training inputs but
for only 4 of 113 evaluation inputs. After \rftv{1} was trained, an LLM agent read all 1,144 unique traces of this set and
of the next round end to end and removed 16 (1.4\%): seven contained contradictions, five drifted off the task, three
dithered and one asserted an unjustified rule. The cleaned set (1,022 rows) is the base of every later training set.

\begin{algorithm*}[t]
\caption{Rejection-sampled trace data (\rftv{1}; bracketed lines are the extensions of later sets).}
\label{alg:rft}
\begin{algorithmic}[1]
\Require labelled inputs $P$ (ARC-AGI-2 training, public evaluation minus holdout $H$); generator $G$ (\rftv{1}: base
\qbig{}; later sets: \rftv{1}); thinking cap $C$; attempts $A$
\For{each input $x \in P$}
  \State [E3: for each boundary input (solve rate $\le 1/2$) and each view $v$ = D8 element $\times$ colour permutation
  fixing 0]
  \For{$a = 1, \dots, A$} \Comment{\rftv{1}: $A=2$, $C=48$K; \rftv{3}: $A\le 8$, stop after two correct}
    \State seed $\gets$ \textsc{sha256}(task, test index, $a$, tag)
    \State $t \gets G(\text{prompt}(x);\ T{=}0.6, p{=}0.95, k{=}20,\ \text{stop at \texttt{</think>}, cap } C)$
    \If{$t$ reached the cap} append the forced-close sentence and \texttt{</think>} \EndIf
    \State $y \gets G(\text{prompt}(x) \Vert t;\ \text{grid grammar},\ \le 8{,}192 \text{ tokens})$
    \State [\rftv{2}+: program $\gets$ $G(\dots \Vert \text{program cue})$; run it on all train pairs and the test input]
  \EndFor
\EndFor
\State keep $(t, y)$ iff $y = $ label $\wedge$ natural completion $\wedge$ task $\notin H$ $\wedge$ no give-up phrase
$\wedge$ no repeated lines $\wedge$ zlib ratio $\ge 0.18$
\State [\rftv{2}+: last grid in $t$ equals $y$, $|t| \le 32$K (E3: 48K); \rftv{3}+: drop lucky hits on small or rarely solved
inputs unless the trace's own program verifies or two attempts were correct]
\State per input keep the $k$ shortest ($k=1$ training, $k=2$ evaluation)
\State [after \rftv{1}: an LLM agent reads every trace of the first two rounds; 16 of 1,144 removed]
\State supervise $t \Vert \texttt{</think>} \Vert y\ [\Vert \text{program}] \Vert \texttt{<|im\_end|>}$ with the prompt masked
\end{algorithmic}
\end{algorithm*}

\subsection{Training recipe}
\label{sec:qwen-train}

We train a LoRA adapter \citep{hu2022lora} of rank 32 and $\alpha=32$ (233.5M parameters) on twelve projection types of the
hybrid architecture: query, key, value and output of the 16 full-attention layers, the five input and output projections
of the 48 Gated-DeltaNet layers \citep{yang2025gdn}, and the gate, up and down projections of all 64 MLP blocks. The base
stays frozen in BF16. We use AdamW with learning rate $10^{-4}$, cosine decay after a 5\% warm-up, gradient clipping at 1.0,
one epoch, micro-batch 1 without packing (packing would leak the recurrent state of the linear-attention layers across
examples) and gradient accumulation 8. The loss is the token mean of cross-entropy over supervised tokens; the prompt is
masked and asserted to be token-identical to the inference prompt. Examples up to 81,920 tokens fit on one 96~GB GPU with
re-entrant gradient checkpointing, activation offload to host memory, and MLPs chunked over 8,192 positions. \rftv{1}
trained 1,033 rows (three exceeded the length limit) in 130 optimizer steps and 7.06 hours; the loss stayed near 0.20
nats per token. After merging, we re-quantized the model to NVFP4 with the recipe of the public \texttt{unsloth}
checkpoint: NVFP4 weights with group size 16 on the MLPs of layers 0 to 55; FP8 weights and activations on attention, the
query-key-value, gate and output projections of the Gated-DeltaNet layers, the output head and the last eight MLPs; BF16 for
the remaining small projections and norms; and FP8 key-value scales, calibrated on 256 training-task samples. With the same
seeds at a 32K cap, the BF16 and NVFP4 models solved 50 and 51 of 172 inputs (grid or verified program); parity at 64K was
not established.

\subsection{Data variants after \rftv{1}}
\label{sec:qwen-variants}

\Cref{tab:qwen-data} lists the training sets. Each later set changed one aspect of the data. \emph{More of the same}
(main2) added second-pass traces at an 88K cap on the 127 inputs of partly unsolved tasks. \emph{Short traces} (short16k)
kept only traces of at most 16K thinking tokens. \emph{Explanations} put verified rule explanations in the thinking
channel (v3b: 284 explanations of official tasks written by an LLM and checked against the labels, plus 1,000
template-generated explanations of synthetic tasks) or in the ordinary answer channel (ft64k: 1,602 explanations, 1,584 of
them of synthetic tasks; \cref{sec:qwen-synth}).
\emph{Program targets} (\rftv{2}) added 777 traces generated by \rftv{1} with the production request; 755 of them carry a
Python program target: the trace's own verified program, a program the model wrote after its own frozen correct thinking
(832 of 929 inputs verified), or a compact program from a library of 1,115 programs that an LLM agent wrote for official tasks and that we checked
in a sandbox. \emph{Hard sampling}
(\rftv{3}) drew up to eight attempts on unsolved inputs, dropped lucky hits on small or rarely solved inputs, and capped new
rows at two per input and per task. \emph{Boundary $\times$ views} (E3) sampled the 300 training inputs that \rftv{1} solves
at most half the time and the 90 T inputs, each in two views (a D8 element and a colour permutation that keeps black fixed,
applied to every grid of the task), at a 48K cap; this set was folded into FINAL and never trained on its own.
\emph{Teacher traces} (FINAL) added 452 traces from external models that return their reasoning text: dots3 (165 rows),
DeepSeek V4.1-Flash (152), two free models of a coding command-line tool (121) and others (14).

Two findings about teacher traces shaped FINAL. First, supplying the known test output as a hint (``never mention it'')
leaks into the reasoning: dots3 solved 8 of 16 previously unsolved inputs this way, and all 8 correct traces mention the
hint. A regular-expression filter for such references also fires on 9\% to 21\% of unhinted correct traces, so we stopped
hinting; one hinted trace passed the filter and is in FINAL. Second, a \emph{leave-one-out view} gives the teacher the
answer legitimately: the real test pairs become extra demonstrations and one training pair, chosen by seed, becomes the
query. In a pilot the teacher solved 8 of 12 such queries and all 8 passed the filters. The FINAL set (1,824 rows, 36.59M
thinking tokens) combined the cleaned \rftv{1} data, the E3 rows, the 452 teacher rows and 35 further self-traces, and
for the first time included the H tasks (74 rows), so the public evaluation set could no longer validate it.

Before FINAL, we measured how well \rftv{1} fits its own data (E1). Comparing the base model and \rftv{1}, the per-token
negative log-likelihood of the thinking fell by 0.047 nats on training-family rows (0.230 to 0.184) but only by 0.020 on
evaluation-family rows (0.294 to 0.274); the difference of the two reductions is 0.027 (bootstrap 95\% interval 0.024 to
0.030). \rftv{1} therefore fits the long evaluation-level traces less than the short training-level ones. A run with a
higher learning rate that was prepared to address this (E2) was cancelled in favour of a single final run.

\begin{table*}[t]
\centering
\caption{Fine-tuning sets for \qbig{}. Rows are training examples (rows actually trained in parentheses), eval rows are
rows from public evaluation tasks (tasks in parentheses), and thinking tokens are total / median / 90th percentile in Qwen
tokens. All sets except the answer-only one contain only traces whose final grid equals the label.}
\label{tab:qwen-data}
\small
\begin{tabularx}{\linewidth}{@{}l>{\raggedright\arraybackslash}Xrrr@{}}
\toprule
Set & Composition & Rows & Eval rows & Thinking tokens \\
\midrule
Answer-only & synthetic + official grids, thinking off & 37,201 & 125 & 0 \\
\rftv{1} & base \texttt{xhigh} self-traces & 1,036 (1,033) & 136 (61) & 17.08M / 12.2K / 38.0K \\
main2 & \rftv{1} + second-pass traces at 88K & 1,109 (1,103) & 152 & 19.66M / 13.0K / 41.2K \\
short16k & traces of $\le$16K tokens & 617 & 4 & 4.07M / 5.7K / 13.4K \\
v3b & 769 own traces + 1,284 explanations (1,000 synthetic) in the thinking channel & 2,053 & 0 & 25.1M \\
ft64k & 1,039 own traces + 1,602 explanations (1,584 synthetic) as answer text & 2,641 & -- & 20.2M + 17.9M \\
\rftv{2} & \rftv{1} + 777 \rftv{1} traces, 755 with programs & 1,813 (1,810) & 152 & 26.08M / 10.7K / 31.0K \\
\rftv{3} & \rftv{2} + hard sampling, rare-hit rule & 1,972 (1,969) & 192 (62) & 29.76M / 12.1K / 30.5K \\
E3 & cleaned \rftv{1} + 315 boundary $\times$ view traces & 1,337 (not trained) & 197 & 23.33M / 13.6K / 38.9K \\
FINAL & E3 + 452 teacher (28 synthetic) + 35 self-traces, H included & 1,824 (1,801) & 414 (106) & 36.59M / 17.2K / 41.8K \\
\bottomrule
\end{tabularx}
\end{table*}

\subsection{Synthetic tasks in the fine-tuning data}
\label{sec:qwen-synth}

Synthetic tasks entered three of the training sets of \qbig{} (\cref{tab:qwen-synth}): 2,231 distinct tasks in total,
1,435 from the 95,993-task corpus of \cref{sec:rules} and 796 from the 72-program bank of the shape-balanced generator.
The largest addition, \emph{synthetic1000}, was built to cover every grid size. It has 1,000 tasks with one test input
each, and every ordered size of the test input from $1\times1$ to $30\times30$ (900 sizes) occurs in at least one task.
205 of its tasks keep the five demonstrations and the test input of a corpus task (205 rules from all eight corpus
families), and 795 are new tasks of 68 of the 72 bank programs, which come in two families: \emph{spatial logic} (a cell
selector, a relation, a symmetry and an occupancy gate; 684 tasks) and \emph{shape sequence} (an order on the cells
followed by a relation, a mask or a layout; 111 tasks). The generating programs replayed all 4,410 pairs, no task equals a
task of the existing fine-tuning data after canonicalisation, and the set takes 8.85M tokens in the training format (at
most 22,022 per task). Every task carries an explanation of its rule that is computed from the verified application of the
program, not written by a model; \cref{tab:rule-examples} gives one rule of each family in the words of these
explanations.

We used the synthetic tasks as supervised explanations and not as self-traces, because the model rarely solved them: in
the trace generation of \rftv{2}, \rftv{1} solved 5 of 45 inputs of synthetic1000 and none of 28 corpus inputs, all 28 of
which reached the 32K cap. v3b put the 1,000 explanations of synthetic1000 in the thinking channel; ft64k put 1,584
explanations of synthetic tasks (1,248 from v2 rules, 164 from v1 rules and 172 from the bank) in the ordinary answer
channel, next to 1,039 self-traces in the thinking channel; and FINAL contains 28 teacher traces of synthetic tasks. No
set with synthetic tasks was better than \rftv{1} on the public evaluation set (\cref{tab:qwen-models}), and the deployed
\rftv{1} contains none. A held-out set of 30 synthetic tasks, three from each of the ten families, shows how far the
models are from the corpus: \rftv{3} solved 1 at a 32K cap and 5 at 64K, and FINAL 1 and 3.

\begin{table*}[t]
\centering
\caption{Synthetic tasks in the fine-tuning data of \qbig{}. ``Corpus'' tasks come from the 95,993-task export of the rule
corpus, ``bank'' tasks from the 72-program shape bank. The result is inputs solved of 172 by one 32K attempt, against 51
for \rftv{1} in the same harness; $^*$2-request harness, against 37 for \rftv{1}.}
\label{tab:qwen-synth}
\footnotesize
\begin{tabularx}{\linewidth}{@{}l>{\raggedright\arraybackslash}Xrrr>{\raggedright\arraybackslash}Xr@{}}
\toprule
Set & Channel of the synthetic rows & Synthetic & Corpus & Bank & Other rows & Result \\
\midrule
v3b & thinking (explanations) & 1,000 & 205 & 795 & 284 official explanations, 769 self-traces & 9$^*$ \\
ft64k & answer text (explanations) & 1,584 & 1,412 & 172 & 18 official explanations, 1,039 self-traces & 43 \\
FINAL & thinking (teacher traces) & 28 & 10 & 18 & 1,796 traces of official tasks & 44 \\
\midrule
All & & 2,231 & 1,435 & 796 & & \\
\bottomrule
\end{tabularx}
\end{table*}

\subsection{Inference procedure}
\label{sec:qwen-infer}

The configuration that reached our best \qbig{} score (44.03) is given in \cref{alg:qinfer}. It runs \rftv{1} in NVFP4 on
four L4 GPUs with tensor parallelism 4, an FP8 key-value cache, three-token MTP, a server limit of 36 sequences and at
most 32 client requests in flight. Compared with
the first \rftv{1} run on the semi-private set (28.19), it also switched from FP8 to NVFP4 weights, added two decode speed-ups (batch-sharded
sampling, $+5.5\%$, and local-argmax draft reduction, $+3.9\%$) and added repair passes, and it made three changes to how
attempts are made:
\begin{enumerate}
\item \emph{One request per attempt with the thinking cap enforced inside the engine.} We patched the engine's thinking
budget so that at the cap it emits the forced-close sentence and \texttt{</think>}, and the grid grammar applies only
after \texttt{</think>}. Natural completions already ended in one request; for attempts that reach the cap, this removes the
second request and its re-prefill. In a 40-task A/B on the four L4 GPUs with the same seeds and a 32K cap, completed attempts
per hour rose from 31.6 to 43.7 (+38\%) and the prefill tokens on the server fell from 3.50M to 0.79M; both arms solved 20
of 59 inputs (33.9\% input-weighted), although the arm without the change completed only 91 of its 118 attempts within the
three-hour test.
\item \emph{A program in the same request.} After the grid, a grammar-forced cue asks for the transformation as a Python
function \texttt{transform(grid)}; the program is executed on all training pairs, and a program that reproduces them all
(``train-perfect'') ranks before the direct grid.
\item \emph{A budget schedule.} Attempt 1 with a 56K cap is started for every input before any second attempt is started;
the queue does not wait for first attempts to finish, so second attempts start while up to 31 first attempts are still
running. Attempt 2 uses a 32K cap on a view with reordered demonstrations; each request's cap shrinks so that it can finish before the deadline at
the measured rate of 15 tokens per second per stream.
\end{enumerate}

\paragraph{Programs after the answer.} \qbig{} does not write programs on its own: none of its 344 \texttt{xhigh} attempts
on the public evaluation set contains Python. The cue after the grid asks for ``The transformation above, written as a
self-contained Python function'' and opens a code block with \texttt{def transform(grid):}; the model writes the body under a
grammar that ends at the closing fence, and the program runs in an isolated interpreter (10~s CPU, 1~GiB memory) on every
training input and on the test input. Such a program checks the answer against the demonstrations and avoids copying
errors in long grids. On the saved base-model attempts (64K cap, two attempts), 98 of 344 programs were train-perfect and
76 of those were right on the test input (77.6\%); using the program as the second answer raised the task score from 38.47
(attempt 1 alone) to 48.47, and ranking programs first over both attempts gave 54.03. A program in a second request lowered
the semi-private score from 26.81 to 21.39, because each request re-read about 55K tokens of context, about 1.8 of the 12 hours. In the same
request, with \rftv{1} at a 32K cap, 37 of 172 programs were train-perfect and 27 of those were right (73\%); ranking them
first rescued 9 inputs and lost none (41 to 50 inputs, task score 25.69 to 31.53) for 4.5\% of the generated tokens, and
consecutive versions of the configuration without and with programs scored 33.61 and 37.22 on the semi-private set. A repair turn that shows the model the
cell-level differences of its best program on each training pair solved 8 more inputs and lost 1 at a 16K budget, but a
replay of the deployed queue indicates that repairs did not run on the hidden set.

\begin{algorithm*}[t]
\caption{Inference and answer selection for \qbig{} on four L4 GPUs (configuration of the 44.03 run).}
\label{alg:qinfer}
\begin{algorithmic}[1]
\Require tasks sorted by prompt length; time limit $L = 12$ h; hard stop $L - 7$ min; rate $r = 15$ tokens/s per stream
\State queue levels: L0 attempt 1 for every input (plain view, cap 56K); L1 attempt 2 (demonstrations shuffled, cap 32K);
L2 repair turn with per-pair program feedback (24K, then 16K unless a program is now train-perfect), queued after both
attempts only if no program is train-perfect and a plain-view program reproduces at least one demonstration, started only
at least 30 min before the hard stop; L3 one extra attempt, only if no program is train-perfect and the two grids differ,
with at least 35 min left
\While{queue not empty and time $< L - 15$ min}
  \State pop the lowest level; cap $\gets \min(\text{cap},\ r\,(\text{hard stop} - \text{now} - 240\,\text{s}) - 3000)$;
  skip if cap $< 4000$
  \State one request: thinking $\to$ forced completion at cap $\to$ grid (grammar) $\to$ \texttt{def transform(grid):} body
  \State run the program on all training pairs and the test input in a sandbox (10 s CPU, 1 GiB)
  \State rewrite the answer file
\EndWhile
\State per input: candidates $\gets$ train-perfect program outputs $\cup$ direct grids
\State order: candidates with $\ge 2$ votes \citep[as in][]{wang2023selfconsistency}, then program outputs by votes,
then grids by votes
\State attempt\_1, attempt\_2 $\gets$ first two distinct candidates (fill with \texttt{[[0]]})
\end{algorithmic}
\end{algorithm*}

\subsection{Validation without the time limit and on the semi-private set}
\label{sec:qwen-eval}

\Cref{tab:qwen-val} puts the offline results of the models next to their semi-private scores where these exist. Offline runs are on the 172
public evaluation inputs on one RTX~PRO~6000 (Colab), with no limit on time; runs on the semi-private set are on four L4
GPUs for twelve hours.
Without the time limit, the base model reached task scores of 49.86 with two 64K attempts, 53.33 with an 88K and a 64K
attempt, and 54.03 with a program after each attempt, and \rftv{1} solved 99 of the 172 inputs (57.6\% of inputs, task
score 59.6) in at least one of its attempts (8 to 22 per input, 13.4 on average). On the semi-private set, the same token budgets did not fit: two 88K attempts scored 18.89 and a program in a
second request 21.39, because many hidden inputs received no answer. Once the schedule fitted the twelve hours, validation
and semi-private scores agreed. For \rftv{1} with a 64K and a 32K attempt, the offline task score was 44.5 on all tasks and 42.5 on
the subset H (mean over four pairings, see \cref{tab:qwen-val}), and the run on the semi-private set with a 56K and a 32K
attempt scored 44.03. These results suggest that throughput and token allocation constrain the deployed score. They do
not isolate how much of the difference between the 54 points reached offline without a time limit and the 44 points on
the semi-private set the token budget causes: the two sets differ in tasks and in training exposure, the offline runs used
another GPU, engine and quantisation, and the offline 64K and 32K combination differs from the deployed 56K and 32K
schedule. The inference procedure of \cref{sec:qwen-infer} is our means of spending the budget.

\begin{table*}[t]
\centering
\caption{\qbig{} models offline without a time limit (public evaluation set, 172 inputs, Colab) and on the semi-private
set (4$\times$L4, 12 hours). Offline rows give inputs solved and the task score; ``grid or program'' counts an
input as solved if the direct grid or a train-perfect program output is right. H is the development subset of 45 inputs.
$^a$Mean over four pairings of one 64K attempt with four separate 32K draws (74, 73, 71 and 68 inputs), each pair ranked by
the production rule; the 32K draws differ in weights (NVFP4 or BF16), GPU and seed, and none uses the reordered
demonstrations of the deployed second attempt. $^c$Half of the tasks answered by design.
$^d$Trained on H and on all evaluation tasks, so its offline numbers are in-sample.}
\label{tab:qwen-val}
\footnotesize
\begin{tabularx}{\linewidth}{@{}l>{\raggedright\arraybackslash}Xrrr>{\raggedright\arraybackslash}Xr@{}}
\toprule
Model & Offline setting & Inputs & Task & H & Semi-private setting & \lb{} \\
\midrule
Base & two 64K attempts, $T=1.0$ & 84 & 49.86 & -- & two 64K attempts & 26.81 \\
Base & 88K and 64K attempts & 89 & 53.33 & -- & two 88K attempts & 18.89 \\
Base & two 64K attempts, program after each & -- & 54.03 & -- & program in a second request & 21.39 \\
\rftv{1} & one 32K attempt, grid or program & 51 & 30.69 & 10 & two 32K attempts, programs & 37.22 \\
\rftv{1} & one 48K attempt, grid or program & 56 & 33.06 & 11 & 48K and 32K attempts & 37.08 \\
\rftv{1} & one 64K attempt, grid or program & 62 & 39.03 & 15 & -- & -- \\
\rftv{1} & 64K and 32K attempts$^a$ & 71.5 & 44.5 & 16.0 & 56K and 32K attempts & \textbf{44.03} \\
\rftv{1} & any of 8 to 22 attempts & 99 & 59.6 & 21 & -- & -- \\
\rftv{2} & one 32K attempt, grid or program & 49 & 30.69 & 12 & 48K on half of the tasks$^c$ & 22.50 \\
\rftv{3} & one 64K attempt, grid or program & 63 & 38.89 & 11 & -- & -- \\
FINAL$^d$ & one 64K attempt, grid or program & 60 & 35.14 & 12 & as for 44.03 & 29.03 \\
\bottomrule
\end{tabularx}
\end{table*}

\Cref{tab:qwen-models} compares the models at fixed thinking caps with the same seeds where possible. At a 64K cap and a
single attempt in the 2-request harness, \rftv{1} solved 63 of 172 inputs and the base model 57 at $T=0.6$; \rftv{1} won 29
inputs and lost 23 (exact sign test $p=0.49$). The base model's single attempts at $T=1.0$ solved 59 and 64 inputs, so the
difference is inside the base model's own variation. On the 90 T inputs, whose correct traces were in the training data,
both solved 46. Mean thinking length at 64K did not change (51.2K and 51.8K tokens). \rftv{1} therefore did not measurably
change single-attempt accuracy at a matched budget. In the production harness at 32K, no later training set was better than
\rftv{1}: \rftv{2} 49 and \rftv{3} 42 inputs against 51, ft64k 43 and FINAL 44 (main2, short16k and v3b were only run in the
2-request harness, where main2 equalled the base model at 64K and short16k solved 34 against 37 at 32K). On the
subset H, \rftv{2} solved 12 against 10 for \rftv{1} at 32K and 9 against 11 at 48K, so no variant was better on both
budgets. With the 2-request harness, v3b solved 9 inputs against 37 for \rftv{1}; 149 of its 172 attempts begin with
``Solution summary.'' and none of them is correct. This suggests that explanations written with knowledge of the answer
teach the model to state a rule without exploring, although v3b also differed in trace count, maximum length and batch
size.

\begin{table*}[t]
\centering
\caption{Single-attempt accuracy of \qbig{} models on the public evaluation set at fixed thinking caps (inputs solved, of
45 for the subset H and of 172 overall). The 2-request harness thinks and then answers in a separate request (FP8
weights for the base model, BF16 for fine-tuned models); the 1-request harness is the production request of
\cref{alg:qinfer} (NVFP4 weights). ``Grid'' counts direct answers; ``two-slot'' also counts inputs whose train-perfect
program output is right. FINAL was trained on H tasks.}
\label{tab:qwen-models}
\small
\begin{tabular}{@{}lllccc@{}}
\toprule
Model & Harness & Metric & 32K & 48K & 64K \\
\midrule
Base FP8, $T=0.6$ & 2-request & grid & -- & -- & 10 / 57 \\
Base FP8, $T=1.0$ (two attempts) & 2-request & grid & -- & -- & 13--17 / 59--64 \\
\rftv{1} & 2-request & grid & 8 / 37 & -- & 13 / 63 \\
v3b & 2-request & grid & 2 / 9 & -- & -- \\
\midrule
\rftv{1} & 1-request & grid & 8 / 40 & 9 / 50 & 14 / 54 \\
\rftv{1} & 1-request & two-slot & 10 / 51 & 11 / 56 & 15 / 62 \\
\rftv{2} & 1-request & two-slot & 12 / 49 & 9 / 54 & -- \\
\rftv{3} & 1-request & two-slot & 7 / 42 & 12 / 56 & 11 / 63 \\
ft64k & 1-request & two-slot & -- / 43 & -- & -- \\
FINAL & 1-request & two-slot & 11 / 44 & -- & 12 / 60 \\
\bottomrule
\end{tabular}
\end{table*}

The semi-private set agrees with the one matched model comparison we made there. FINAL and \rftv{1}, run with identical
code and container image, scored 29.03 and 44.03. With a per-draw standard deviation of 1.94 points, measured on
repeated \nvarc{} runs on the semi-private set (2.7 for the difference of two draws), a 15-point drop is far larger than the run-to-run
variation. We did not test why. Two candidate causes are that a quarter of FINAL's rows come from other models, whose
reasoning style differs from the native one, and that FINAL completes its thinking later: in the production harness at 64K,
20 of its natural completions fall between 56K and 64K tokens against 9 for \rftv{1}, so a 56K cap forces more
completions.

%% file: sec/flashnext.tex
\section{Pruning the experts of \qflash{} to fit four L4 GPUs}
\label{sec:flash}

The third line of work used \qflash{} \citep{qiu2026qwen38next}, a sparse mixture-of-experts reasoning model from the same
family as \qbig{}. We did not fine-tune it. Offline it was as accurate per attempt as \qbig{} and gained more from long
thinking, but on four L4 GPUs it first generated about a third of the tokens per second of \qbig{}, because its weights left
little memory for the key-value cache. This section describes how we ran it on the evaluation hardware (four L4 GPUs), how we chose the
experts to remove, and what the removal changed in accuracy, memory and tokens.

\subsection{Model and serving}
\label{sec:flash-model}

\qflash{} has 48 layers with hidden size 2,560: 36 Gated DeltaNet layers \citep{yang2025gdn} and, in every fourth layer, full
attention with a sparse-attention indexer (compression ratio 4, budget 2,048 positions). Every layer has 512 routed experts
with intermediate size 640, of which the router selects 10 per token, and one gated shared expert as in
\citet{dai2024deepseekmoe}. A multi-token-prediction layer, itself a mixture of 512 experts, serves as the draft model. The
model has 125B parameters, of which 6B are active per token, plus a 51B-parameter n-gram embedding table that is kept off
the accelerator \citep{qiu2026qwen38next}; by our count from the configuration, the routed experts of the 48 main layers
hold 120.8B parameters.

We served a community W4A16 checkpoint: routed experts in symmetric INT4 with groups of 128, the Gated DeltaNet projections
in per-channel INT8, the n-gram table in FP8 and the remaining weights in BF16. It occupies 115.22~GiB, of which the n-gram
table (47.7~GiB) stays in host memory. The serving engine is Merlin, a community fork of vLLM \citep{kwon2023vllm} for
Ampere and Ada GPUs, installed from prebuilt wheels: version 2.5.2 (based on vLLM 0.30) at first and version 3.0
(based on vLLM 0.32) later. It splits the model into two pipeline stages of 25 and 23 layers, each with tensor
parallelism 2 and expert parallelism, and stores the key-value cache in FP8 with calibrated static scales; against a BF16
cache, the FP8 cache changed the teacher-forced log-likelihood by 0.0002 to 0.0004 nats per token in each of five length
bins. Speculative decoding uses the multi-token-prediction layer with four draft tokens and an INT8 output head for the
drafts.

\subsection{Difficulties on four L4 GPUs}
\label{sec:flash-l4}

\paragraph{Memory.} An L4 exposes 22.03~GiB to PyTorch. With all 512 experts, the weights on each GPU took 16.9 to
17.3~GiB, which left about 2.2~GiB per GPU for activations, CUDA graphs and the key-value cache. \Cref{tab:flash-serving}
lists the layouts we ran with the unpruned model. Stock vLLM with tensor parallelism 4 and expert parallelism fitted 94,001
cache tokens, enough for four concurrent requests, and generated 164 tokens per second; the server was ready 68 minutes
after the run started. Tensor parallelism 4 or a four-stage pipeline in Merlin fitted 96,743 to 255,273 tokens at 136 to
168 tokens per second. Two pipeline stages with tensor parallelism 2 and an FP8 cache fitted 420,026 tokens and reached
about 250 tokens per second, and Merlin 3 with an INT8 draft head 547,542 tokens and 342 tokens per second. Even then the
engine ran only 13 to 19 requests at a time, since one attempt that thinks for 96K tokens occupies about a fifth of a
547,542-token cache.

\begin{table*}[t]
\centering
\caption{Serving the unpruned \qflash{} (512 experts per layer) on four L4 GPUs. TP and PP are the tensor- and
pipeline-parallel degrees; all layouts use expert parallelism. Throughput is total generated tokens per second, measured as
noted: over the whole run, as the steady per-minute rate after the first five minutes, or over the first 150 ten-second
windows of a probe; rows with different definitions are not directly comparable.}
\label{tab:flash-serving}
\footnotesize
\begin{tabularx}{\linewidth}{@{}>{\raggedright\arraybackslash}Xrrl@{}}
\toprule
Engine, layout and cache & Cache tokens & Running & Tokens/s \\
\midrule
stock vLLM 0.30, TP4, BF16 cache & 94,001 & 4 & 164 (whole run) \\
Merlin 2.5.2, TP4, BF16 cache & 96,743 & -- & 137 (steady) \\
Merlin 2.5.2, TP4, FP8 cache & 170,393 & -- & 136--168 (steady) \\
Merlin 2.5.2, PP4, FP8 cache & 255,273 & -- & 136 (steady) \\
Merlin 2.5.2, TP2 $\times$ PP2, FP8 cache & 420,026 & 12.6 & 249--256 (steady) \\
\quad + four draft tokens, no prefix caching, 96K context & 466,151 & 16.7 & 301 (first 150 windows) \\
Merlin 3.0, TP2 $\times$ PP2, FP8 cache, INT8 draft head & 547,542 & 18.9 & 342 (first 150 windows) \\
\bottomrule
\end{tabularx}
\end{table*}

\paragraph{Code paths and checkpoints.} Stock vLLM rejected the quantised n-gram embedding of the first W4A16 checkpoint we
tried, and Merlin's host-memory offload of the n-gram table crashed with two other published checkpoints because the
offload process initialised CUDA; one checkpoint booted on every layout, and all runs below on the evaluation hardware use it. Merlin's custom
all-reduce failed on the PCIe-connected L4s and was disabled.

\paragraph{Speculative decoding.} The indexer's cache ring must divide the 4,096-token attention block, so only 1 to 4 or 9
to 12 draft tokens are valid; six draft tokens did not boot, and SGLang \citep{zheng2024sglang} refuses more than four. We
used four; three were 1.2\% slower on the 384-expert model. SGLang itself, without pipeline parallelism, fitted about
434,000 cache tokens and generated 221 to 255 tokens per second in the probes where Merlin generated 639 to 657.

\paragraph{Graph capture and start-up.} Capturing CUDA graphs ran out of memory on the last pipeline stage for several cache
sizes, so every configuration needed its own largest cache size that boots. Merlin 3 was ready 635 to 1,033 seconds after
the run started, including a 7.6~GB copy of its Python environment and, when the system pruned experts at start-up,
109 to 287 seconds to drop them.

\paragraph{Long requests and the time limit.} At 9 to 11 tokens per second per stream, a 96K attempt runs for close to three
hours. Our runner had a 7,200-second read timeout with three retries, so every attempt still running two hours after it
started was aborted and sent again from the beginning; this affected the first two Merlin runs (25.00 and 32.64).
With the read timeout removed, a three-hour probe completed 63 attempts without error, 8 of them 7,382 to 8,541 seconds
after they started. A run with an internal limit of 12 hours and 15 minutes was stopped for exceeding the time
limit and returned no score, and later runs stop at 11 hours and 50 minutes.

\subsection{Accuracy before pruning}
\label{sec:flash-before}

Offline comparisons ran on one RTX~PRO~6000 with stock vLLM, a BF16 cache, effort \texttt{xhigh}, $T=0.6$, top-$p$ 0.95 and
top-$k$ 20, on a checkpoint with the same INT4 experts but BF16 attention and Gated DeltaNet weights. We used two panels of
public evaluation inputs: the dev panel, 82 inputs of 59 tasks (the 45 inputs of the subset H and 37 inputs never
solved by the \qbig{} line), and the fresh panel, 90 inputs of the other 61 tasks. \Cref{tab:flash-before} compares the
unpruned model with \qbig{}. On the full public evaluation set with two 64K attempts, the two models tied (task scores
49.31 and 49.86, 52 fully solved tasks each). On the dev panel at 48K, \qflash{} tied \rftv{1}: 29 of 164 attempts each,
with mean thinking lengths of 44.5K and 44.3K tokens. At 96K, three seeds of \qflash{} solved 30, 23 and 25 of 82 inputs
(mean 26.0) against 13 for one seed of \rftv{1}; a bootstrap over the 59 tasks gives a difference of $+13.0$ inputs (95\%
interval $+4.7$ to $+21.3$). Only the best seed, 30, used the same request seeds as the \rftv{1} run; in this selected
pair, 23 inputs were solved only by \qflash{} and 6 only by \rftv{1} (sign test $p=0.0023$, not adjusted for the choice
of seed). From 48K to 96K the solves of \qflash{} in that seed rose from 16 to 30, while those of \rftv{1} stayed at 13.

\begin{table*}[t]
\centering
\caption{The unpruned \qflash{} against \qbig{} offline. The full-evaluation run of \qflash{} used INT8 attention and Gated
DeltaNet weights, 12 concurrent requests and a BF16 cache; the base \qbig{} used FP8 weights, an FP8 cache and 32
concurrent requests. Dev-panel runs compare against \rftv{1} in NVFP4.}
\label{tab:flash-before}
\footnotesize
\begin{tabularx}{\linewidth}{@{}>{\raggedright\arraybackslash}Xlrr@{}}
\toprule
Set and setting & Metric & \qflash{} & \qbig{} \\
\midrule
Public evaluation, 172 inputs, two 64K attempts & task score & 49.31 & 49.86 \\
 & inputs solved & 80 & 84 \\
 & completion tokens per attempt & 55,057 & 53,756 \\
Dev panel, 48K, two seeds & attempts correct (of 164) & 29 & 29 \\
Dev panel, 96K, one seed & inputs solved (of 82) & 30 & 13 \\
 & thinking tokens per attempt & 68,541 & 71,592 \\
\bottomrule
\end{tabularx}
\end{table*}

On the semi-private set, the first two runs used stock vLLM at effort \texttt{low} with 40K caps and scored 8.61 and 6.11. With
Merlin 3, all 512 experts and caps of 96K and 48K, the score was 25.00, with the timeout fault above. Our simulator of the
twelve-hour run replays offline attempts at the measured throughput; its forecast for the later 53.33 run, made before the
score was known, was 53.0 to 54.5. For the unpruned model at these caps it leaves 160 of about 343 hidden test inputs
without any answer. Accuracy per attempt was therefore not the constraint; tokens were.

\subsection{Choosing the experts to remove}
\label{sec:flash-prune}

Expert pruning removes whole experts with their router rows, so the model keeps its architecture and needs no retraining.
Criteria proposed for it include routing frequency, the reconstruction loss of the layer output
\citep{lu2024notallexperts} and the merging of experts with similar routing \citep{li2024mcsmoe}. REAP \citep{lasby2026reap}
scores expert $j$ of a layer by its router-weighted activation norm over the tokens $\mathcal{X}_j$ routed to it,
\begin{equation}
S_j = \frac{1}{|\mathcal{X}_j|} \sum_{x \in \mathcal{X}_j} g_j(x)\, \lVert f_j(x) \rVert_2 ,
\label{eq:reap}
\end{equation}
where $g_j(x)$ is the gate value and $f_j(x)$ the expert output, and removes the experts with the lowest scores. We removed
the same number of experts from every layer, because the model configuration has one expert count for all layers. No
saliency list covers the multi-token-prediction layer, so there we kept the experts with the largest router-row norm.

\paragraph{Lists.} We did not compute REAP scores ourselves. For 448, 384 and 288 experts per layer we used the lists of
three public REAP builds of the same model, recovered by matching their router rows exactly against the original
checkpoint. Their author reports calibrating the 384 and 288 builds on about 686K tokens of agentic-coding traffic, and
the 288 list is a subset of the 384 list in all 48 layers. To adapt the lists to ARC, we recorded routing on 60 single-input ARC-AGI-2
training tasks: the unpruned model thought for up to 96K tokens (53 tasks solved), and a prefill replay of the 60 traces
recorded 864M routing decisions over 1.80M tokens. For each expert $e$ we computed an ARC routing frequency $F_e$, its share
of the routing decisions, averaged over tasks within eight length bins (prompt, six 16K bins of thinking, answer) and then
weighted towards long thinking, with bin weights from 0.05 for the prompt to 0.19 for 48K to 64K tokens. A second capture
recorded routing on 188,607 program tokens of 236 programs written for 59 training tasks. No evaluation task entered
either capture.

\paragraph{Variants.} Besides the three public lists, we built lists from these statistics. ARC-288 keeps the 288 experts
with the highest $F_e$ in each layer. CORE-320 and CORE-300 keep REAP-288 and add the 32 or 12 experts with the highest
$F_e$ among the 96 that REAP-384 keeps beyond it. s288p exchanges, in each layer, up to 16 experts of REAP-288 that are
nearly unused on ARC for the next REAP-384 experts by $F_e$, and never removes an expert that serves one task or length bin
heavily; 753 experts were exchanged, 15.7 per layer on average. c8 adds to s288p, in each layer, 8 experts of REAP-384 that
program tokens prefer, which gives 296 experts per layer; 6.2 of the 8 were chosen for code on average and the rest by
$F_e$. \Cref{alg:flashprune} gives the two constructions.

\begin{algorithm*}[t]
\caption{Construction of the s288p and c8 expert lists for one layer.}
\label{alg:flashprune}
\begin{algorithmic}[1]
\Require REAP lists $K_{288} \subset K_{384}$; counts $C_{t,e,b}$ of routing decisions of the unpruned model on 60
training traces (trace $t$, expert $e$, length bin $b$) with $N_{t,b}$ tokens; bin weights $w_b$; routing on the program
tokens of 236 programs for 59 training tasks
\State $F_e \gets \sum_b w_b \cdot \operatorname{mean}_{t:\,N_{t,b}>0}\, C_{t,e,b} / (10\,N_{t,b})$
\Comment{10 experts per token}
\State $c_e \gets \sum_{t,b} C_{t,e,b}$;\quad $u \gets$ mean of $c_e$ over the 512 experts
\State specialist$(e) \gets$ some $(t,b)$ has $N_{t,b} \ge 256$ and $C_{t,e,b} > 0.05\,N_{t,b}$
\State weak $\gets$ up to 16 experts of $K_{288}$ with $c_e \le 0.06\,u$ and not specialist, smallest $c_e$ first
\State $K_{\mathrm{s288p}} \gets (K_{288} \setminus \text{weak}) \,\cup$ the $|\text{weak}|$ experts of $K_{384}
\setminus K_{288}$ with the highest $F_e$
\State pool $\gets K_{384} \setminus K_{\mathrm{s288p}}$ (96 experts)
\State code $\gets$ experts of pool that are among the top program-token experts in $\ge 90\%$ of 40 task-level half
samples, pass a within-task permutation test of the maximum $t$ statistic at level 0.05, and receive at least 1.5 times
the uniform share of program-token decisions; keep at most 8, by program-token share
\State $K_{\mathrm{c8}} \gets K_{\mathrm{s288p}} \cup \text{code} \,\cup$ the $8 - |\text{code}|$ experts of pool
$\setminus$ code with the highest $F_e$
\State drop every expert outside the list with its router row; renumber the kept experts in their original order
\end{algorithmic}
\end{algorithm*}

\paragraph{Evaluation.} Lists were compared offline on the dev panel at a 96K cap with three seeds (246 input and seed
pairs) and the same request seeds in every arm, and the last candidates also on the fresh panel (270 pairs). Before testing
s288p and CORE-320, we fixed the rule for adopting a list with fewer experts than REAP-384: the lower end of a one-sided
97.5\% task-bootstrap interval for the difference in task-weighted accuracy above $-1.0$ point, and at least 5\% higher
measured throughput. Speed and cache size were measured on the evaluation hardware with Merlin 3, the INT8 draft head and four draft tokens,
on 60 single-input public evaluation tasks for 6,000 seconds; we report throughput over the first 150 ten-second windows,
while the cache fills, and over the full probe. The lists were built from training tasks only, but we chose among them by
accuracy on the two public evaluation panels, so the panel scores of the adopted lists are optimistic.

\subsection{Results after pruning}
\label{sec:flash-after}

\begin{figure*}[tp]
\centering
\includegraphics[width=\linewidth]{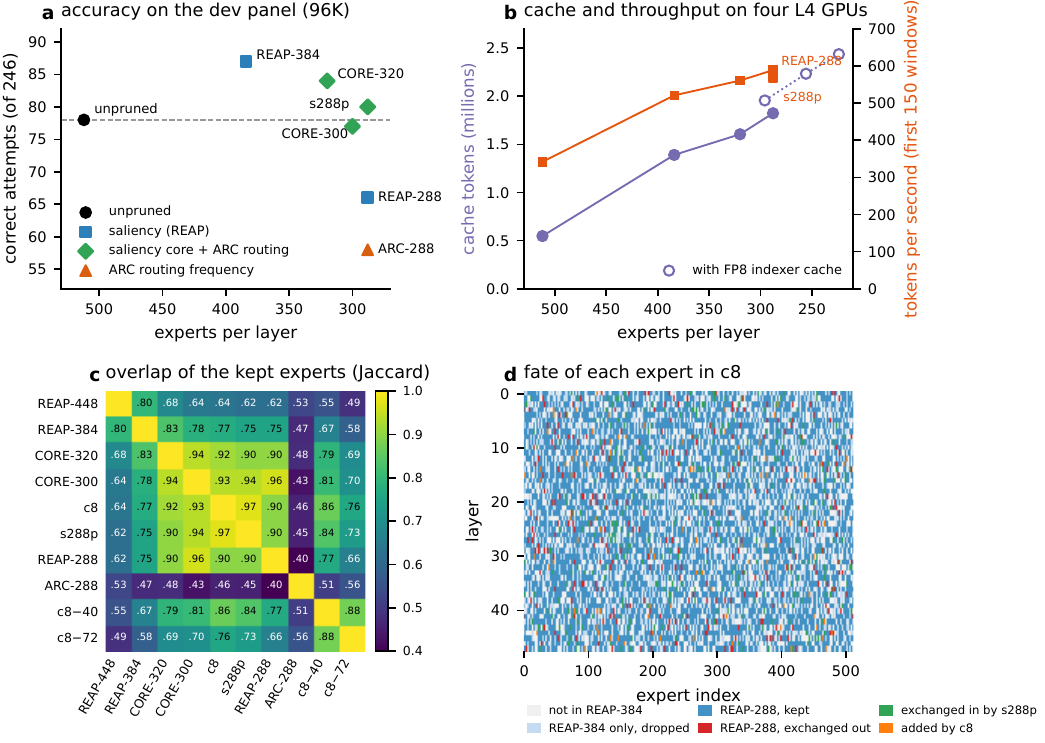}
\caption{Expert lists of \qflash{}. (a)~Correct attempts of 246 on the dev panel at a 96K cap by experts per layer and by
the criterion that built the list (REAP-448, measured on one seed only, is left out); the dashed line is the unpruned
model. (b)~Key-value cache tokens and throughput over the first 150 windows of the L4 probe (\cref{tab:flash-prune});
open circles are lists served with the FP8 indexer cache. (c)~Jaccard overlap of the kept experts, averaged over the 48
layers. (d)~What c8 does with each of the 512 experts of each layer: experts outside REAP-384, experts of REAP-384 that c8
drops, the REAP-288 core that c8 keeps, the 753 REAP-288 experts that s288p exchanged out and the experts it took in
their place, and the 8 experts per layer that c8 adds.}
\label{fig:flash-prune}
\end{figure*}

\Cref{tab:flash-prune} lists the variants and \cref{fig:flash-prune} plots them. Removing a quarter of the experts with REAP-384 lost no accuracy that we could
measure: 87 of 246 attempts against 78 for the unpruned model, at least as many on each seed (34, 26 and 27 against 30, 23
and 25; 31 attempts solved only after pruning and 22 only before, $p=0.27$), with similar rates of natural completion
(70.7\% for REAP-384 and 70.3\% for the unpruned model) and similar thinking lengths (68.5K and 69.3K tokens), pooled over
the three seeds. The 246 attempts are 82 inputs of 59 tasks crossed with three seeds and are not independent, so the
sign test over attempts is descriptive. The weights on each GPU fell from about 17.0 to 13.3 and
13.7~GiB, the cache grew from 547,542 to 1,391,054 tokens, and throughput over the first 150 windows rose from 342 to 521
tokens per second ($+52\%$).

\begin{table*}[t]
\centering
\caption{Expert lists for \qflash{}. Checkpoint size and weights per GPU (pipeline stage 0 / stage 1) are from the serving
logs; the cache pin is the memory reserved for the key-value cache, set to the largest value that booted. Throughput is
total tokens per second on the 60-task L4 probe over the first 150 windows and over the full probe. ``Dev'' counts
correct attempts of 246 on the dev panel at 96K (three seeds). $^a$One seed, of 82. $^b$Two seeds, of 164. $^c$With an
FP8 indexer cache, which by the cache formula of the serving logs holds about 6\% more tokens at equal pin; the measured
gain of 11\% for s288p also includes a larger pin. c8$-$40 and c8$-$72 remove from c8 the experts with the
lowest $F_e$ that are neither code experts nor specialists.}
\label{tab:flash-prune}
\footnotesize
\setlength{\tabcolsep}{4.5pt}
\begin{tabular}{@{}lrrrrrrr@{}}
\toprule
List & Experts & Checkpoint & Weights/GPU & Pin & Cache tokens & Tokens/s & Dev \\
 & per layer & (GiB) & (GiB) & (GiB) & & (150 / full) & \\
\midrule
unpruned & 512 & 115.22 & 16.9 / 17.0 & 2.3 & 547,542 & 342 / -- & 78 \\
REAP-448 & 448 & -- & 15.1 / 15.5 & -- & -- & -- & 26$^a$ \\
\textbf{REAP-384} & 384 & 100.38 & 13.3 / 13.7 & 5.8 & 1,391,054 & 521 / 433 & \textbf{87} \\
CORE-320 & 320 & 92.96 & -- & 6.7 & 1,605,632 & 561 / 463 & 84 \\
CORE-300 & 300 & -- & -- & -- & -- & -- & 77 \\
\textbf{c8} & 296 & 90.17 & -- / 11.2 & 7.7 & 1,957,095$^c$ & -- & -- \\
\textbf{s288p} & 288 & 89.25 & 10.5 / 11.0 & 7.6 & 1,823,909 & 566 / 492 & 80 \\
REAP-288 & 288 & -- & -- & 7.6 & 1,823,909 & 589 / 500 & 66 \\
ARC-288 & 288 & -- & -- & -- & -- & -- & 58 \\
c8$-$40 & 256 & 85.54 & 9.4 / 10.1 & 8.8 & 2,234,566$^c$ & -- & -- \\
c8$-$72 & 224 & -- & 8.5 / 9.2 & 9.6 & 2,438,044$^c$ & -- & -- \\
\bottomrule
\end{tabular}
\end{table*}

Deeper cuts by a single criterion lost accuracy. REAP-288 solved 66 attempts and the frequency list ARC-288 58, the lowest of
all lists (23 attempts gained and 43 lost against the unpruned model, $p=0.019$). The damaged lists complete their thinking
earlier and think less: 85\% of the attempts of ARC-288 were natural completions, with 62.6K thinking tokens on average, against 70.3\% and
69.3K for the unpruned model (all pooled over three seeds). The coding-calibrated REAP lists and the ARC-frequency lists select substantially
different experts; the comparison changes the criterion and the calibration domain together, so it is not a test of
independence. Each REAP list drops about the share of ARC routing decisions that
a random cut of the same size would drop (10.8\%, 24.9\% and 42.3\% for 448, 384 and 288 experts, against 12.5\%, 25\% and
43.8\%), the ARC-frequency top 384 shares 287 experts per layer with REAP-384 where chance gives 288, and among the 296
experts of c8 the rank correlation between REAP saliency on ARC traces and $F_e$ is 0.17 to 0.62 by layer. The lists themselves split the
same way (\cref{fig:flash-prune}c): ARC-288 has a Jaccard overlap of 0.40 to 0.56 with every other list, while the
lists built on the REAP-288 core overlap by 0.90 to 0.97. Keeping the
REAP-288 core and changing only its least-used experts was what worked: s288p solved 80 attempts on the dev panel, and on
the fresh panel it scored 58.38\% task-weighted against 52.00\% for REAP-384 ($+6.38$ points, lower 97.5\% bound $-0.73$),
at 13.6\% higher full-probe throughput (492 against 433 tokens per second). It passed the adoption rule; CORE-320 did not
(lower bound $-4.46$).

Two screens that avoid generation did not replace these tests. The teacher-forced excess negative log-likelihood against the
unpruned model on 49 reference traces ranked the public lists as expected (3.9, 17.4 and 68.5 milli-nats per thinking token
for 448, 384 and 288 experts) but gave ARC-288 20.3, close to REAP-384, so its pre-registered check failed. The mean
probability of the closing tag, a measure of pressure to stop thinking taken in 16K-token windows of the reference traces,
separated them: $3.07\times10^{-4}$ for ARC-288 against $1.68\times10^{-4}$ for REAP-384 and $1.26\times10^{-4}$ for the
unpruned model.

Pruning also changed program writing, which the generation tests on grids did not show. In unmatched runs on the evaluation
hardware, 91 of 248 programs written by s288p did not compile, mostly through indentation errors, against 1 of 144 for
REAP-384; s288p had removed 31.2\% of the routing decisions of program tokens, against 18.0\% for REAP-384. In a matched
replay on Colab, in which both lists continued the same 165 frozen thinking prefixes of s288p, the eight code experts of c8
reduced the share of programs that do not compile from 37.6\% (62 of 165) to 5.5\% (9 of 165) and raised the share of
train-perfect programs from 16.4\% to 29.1\%. The replay measures programs, not accuracy on the hidden set, and its rates
are not comparable with those of the L4 runs (REAP-384 failed to compile 7.3\% of its replay programs). Against
s288p, the production use of programs gained 3.88 points (standard error 2.09), and over 516 dev and fresh pairs thinking
alone lost 1.34 points (standard error 2.44). Cutting c8 further to 256 or 224 experts added 14\% and 25\% to the cache, but
the production mix lost 2.16 and 3.02 points at 272 and 256 experts, and a 256-expert list with every code expert and
specialist protected lost 17.8 points in the agent configuration of \cref{sec:flash-lb}, where its median first-turn
thinking fell from 27,540 to 13,441 tokens. We kept c8.

The main observed benefit was the additional cache capacity. At 56 running requests, decode cycles were similar: in
matched windows (no prompt tokens, exactly 56 running requests) a cycle took about 258.5, 260.6 and 263.9 milliseconds
with 224, 256 and 288 experts, and minutes 10 to 30 of the probes gave 590 to 604 tokens per second for all three lists.
In these probes fewer experts did not make a step noticeably faster; they freed memory, and more requests fit in it.

\subsection{Tokens before and after pruning}
\label{sec:flash-tokens}

\Cref{tab:flash-tokens} summarises the token budget before and after pruning. From stock vLLM with all experts to the
production configuration (c8 with an FP8 indexer cache), the cache grew from 94,001 to 1,957,095 tokens, a factor of 20.8,
and the number of running requests from 4 to 56. Within Merlin 3, pruning to 384 experts multiplied the cache by 2.5
(547,542 to 1,391,054 tokens), and the 296 experts of c8 with an FP8 indexer cache and a larger pin by 3.6 (1,957,095
tokens). Thinking per attempt changed little: pooled over three seeds at a 96K cap, the unpruned model thought for 69.3K
tokens on average and REAP-384 for 68.5K. In production-scale previews on the 120 public evaluation tasks, with the
twelve-hour schedule, the pruned model generated 17.7M thinking tokens in 39,911 seconds with direct answers and 15.6M in
37,230 seconds with the tool agent. Counting all generated tokens (thinking, answers and programs), the server's ten-second
windows after start-up averaged 495 and 469 tokens per second, and 77.8\% of the windows of the agent run were limited by
the cache.

\begin{table*}[t]
\centering
\caption{Tokens before and after expert pruning. Rows not marked as offline or simulated are measured on four L4 GPUs; ``simulated'' rows come from our run
simulator for about 343 hidden test inputs. Offline thinking lengths are on the dev panel at a 96K cap, pooled over three
seeds.}
\label{tab:flash-tokens}
\footnotesize
\begin{tabularx}{\linewidth}{@{}>{\raggedright\arraybackslash}Xrr@{}}
\toprule
Quantity & 512 experts & Pruned \\
\midrule
Cache tokens, stock vLLM / Merlin 3 & 94,001 / 547,542 & 1,957,095 (c8) \\
Running requests on the probe & 4 / 18.9 & 56 \\
Tokens per second, first 150 windows (Merlin 3) & 342 & 521 (REAP-384), 566 (s288p) \\
Tokens per second, full 6,000 s probe & -- & 433 (REAP-384), 492 (s288p) \\
Thinking tokens per attempt, offline, 96K cap & 69.3K & 68.5K (REAP-384) \\
Thinking tokens per twelve-hour preview (120 tasks) & -- & 17.7M direct, 15.6M agent \\
Inputs without any answer, 96K/48K caps (simulated) & 160 & 46 (s288p) \\
Inputs without any answer, 64K/48K caps (simulated) & -- & 2 (s288p) \\
\bottomrule
\end{tabularx}
\end{table*}

Even after pruning, a twelve-hour run is limited by tokens. At caps of 96K and 48K, the simulator leaves 46 of about 343
hidden inputs without an answer and gives almost no input a second attempt; the answer file of the 44.17 run had the
size that this forecast predicts. Most of the remaining tokens go to forced completions. Of 1,032 attempts of s288p at
a 64K cap, 38.7\% reached the cap and ended wrong, and they used 50.3\% of all tokens; natural completions that
were right were 36.9\% of attempts and used 26.2\% of tokens. Of the correct natural completions, 44.6\% came within 32K tokens and 89.0\%
within 56K, so shorter caps lose answers. On the evaluation hardware, 5 of 46 answers forced at the cap were right. The simulator, calibrated
on the 53.33 run, values throughput at about 0.2 points per percent and cache size at about 0.13 points per percent near the
production configuration.

\subsection{Inference changes on the semi-private set}
\label{sec:flash-lb}

\Cref{tab:flash-lb} lists every run of \qflash{} on the semi-private set. Ten scored runs moved the semi-private score
from 25.00 to 56.67, but most steps changed several factors and only the 54.44 configuration was scored twice (54.44 and
55.28), so the steps are not separately
resolved. Pruning to 384 experts and then to s288p, with the timeout fault fixed in the second step, took the score from
25.00 to 32.64 and 44.17. Four inference changes, each tested offline first, entered the 53.33 run together with c8: a
prompt that states the grid sizes and writes each grid row on its own line ($+5.23$ points task-weighted over 516 pairs,
probability of a positive effect 0.979), answer tokens sampled at temperature 0.3 after the thinking (the same grid
accuracy; 29.6\% of programs train-perfect against 22.1\% at 0.6), the transfer of a train-perfect program found
for one test input to the task's other inputs, and caps of 96K for single-input and 64K for multi-input tasks with second
attempts started early. Transferred outputs replace the program outputs of an input that has no train-perfect program of
its own and then vote with its grids; when the input has its own, they do not vote and rank after its grids. Second
attempts are queued when their first attempts finish, ordered by a confidence class and then by thinking length, with
every multi-input task after every single-input task; if the first attempt of a single-input task completed naturally
with a grid in fewer than 40,000 thinking tokens and produced no train-perfect program, its second attempt is placed ahead
of all unstarted first attempts, and the coverage guard that could defer such a promotion was disabled. The last step replaced the first attempt by a tool-using agent (\cref{alg:flash-agent}): the model may call a function that
runs its candidate program on the demonstrations and returns the per-pair errors, at most twice, within a total of 96K
tokens for single-input and 64K for multi-input tasks. Each call is run by two checkers in parallel: the tool checker
(\texttt{check\_transform}), whose per-pair report is returned to the model, and the production checker (the runner's
\texttt{check\_program}), whose result also ranks the candidates of the answer file. The loop ends when both find the program
train-perfect with the same valid test output, after the second check, or when the remaining budget cannot hold another
turn; a final turn then follows unless the second turn has already written the answer. In a 172-input preview of this
configuration, 87 attempts were accepted at the first check and 25 at the second, 53 used both checks without
acceptance, 7 had no room for a second check, none answered in the second turn, and the two checkers never disagreed in
250 checks. On 71 single-input evaluation inputs in two draws, the agent solved
60.6\% against 57.0\% for a direct 96K attempt; in twelve-hour previews on the public evaluation set (input-weighted) it
led direct answers by 56.98 to 52.91 after 6.25 hours and by 60.47 to 59.88 at the end. Unlike the pilot with \qbig{} (\cref{app:failed}), this agent is limited to
checking a program it has already written, and its first turn thinks for up to 80K tokens. Batch-sharded sampling
($+6.4\%$ throughput) and a draft-sampling option of the engine ($+2.1\%$) were added in the same run. A second draw of this configuration scored 55.28, and the same configuration with a 75K agent budget for
single-input tasks (caps of 59,000 and 8,000 tokens instead of 80,000 and 8,000) scored 56.67; the two differ by 1.39
points, within the run-to-run variation.

\begin{algorithm*}[tp]
\caption{First attempt of the 54.44 run: the \qflash{} tool agent for one test input. The 56.67 run gives single-input tasks 75K tokens in total, with caps of 59,000 and 8,000.}
\label{alg:flash-agent}
\begin{algorithmic}[1]
\Require task with demonstrations $D$ and test input $x$; budget $B$ of generated tokens and thinking caps $(c_1, c_2)$:
$B = 96\text{K}$, $(80\text{K}, 8\text{K})$ for tasks with one test input and $B = 64\text{K}$, $(48\text{K}, 8\text{K})$
otherwise; final thinking $c_f = 4\text{K}$; at most 2 checks
\State context $\gets$ the row-wise prompt of $D$ and $x$ with the grid sizes, plus the tool \texttt{check\_transform}
\State $n \gets 0$; $c \gets c_1$; $\mathit{answered} \gets \textbf{false}$
\While{$n < 2$ and the budget leaves at least 1,000 thinking tokens after reserving the call, the tool response and the
final turn}
  \State think for at most $c$ tokens, then emit a \texttt{check\_transform} call; from the second turn on, the final grid
  and program are also allowed
  \If{the turn wrote the final answer} $\mathit{answered} \gets \textbf{true}$; \textbf{break} \EndIf
  \State $n \gets n + 1$ (a call that cannot be parsed also counts and is not run); run the transform on $D$ and $x$ with
  the tool checker (\texttt{check\_transform}) and, in parallel, the production checker (the runner's
  \texttt{check\_program})
  \State append at most 2,500 tokens of feedback from the tool checker: the result of every pair, the expected and
  produced grids of up to two failing pairs, up to 12 wrong cells per pair, and the test output if it has at most 225
  cells; add a note if only the tool checker finds the program train-perfect
  \If{both checkers find the program train-perfect with the same valid test output} \textbf{break} \EndIf
  \State $c \gets c_2$
\EndWhile
\If{\textbf{not} $\mathit{answered}$}
  \State final turn: think for at most $c_f$ tokens (or close at once if no room is left), then write the grid and a
  program under the grid grammar at $T = 0.3$
\EndIf
\State attempt 1 $\gets$ the final grid; checked programs that the production checker finds train-perfect, in check
order, and then the final program if it passes, join the program candidates; other programs that ran with a valid test
output join the weak pool
\State on an infrastructure error that left neither a grid nor a checked program, run the direct attempt instead
\end{algorithmic}
\end{algorithm*}

\begin{table*}[t]
\centering
\caption{All runs of \qflash{} on the semi-private set, in the order they were made. Caps are thinking caps for attempts 1
and 2. Runs from the third row on use Merlin 3, four draft tokens and $T=0.6$ unless noted; ``1-input'' and ``multi'' are
tasks with one or with several test inputs.}
\label{tab:flash-lb}
\footnotesize
\begin{tabularx}{\linewidth}{@{}r>{\raggedright\arraybackslash}Xr@{}}
\toprule
Experts & Configuration (change from the previous row) & \lb{} \\
\midrule
512 & stock vLLM, effort \texttt{low}, 40K & 8.61 \\
512 & same, attempts of 24K and 40K & 6.11 \\
512 & Merlin 3, \texttt{xhigh}, 96K / 48K, client timeout of 7,200 s & 25.00 \\
384 & REAP-384 & 32.64 \\
288 & s288p, timeout removed & 44.17 \\
288 & 64K / 48K, second attempts ordered by confidence, FP8 indexer cache, $T=1.0$ & 40.97 \\
296 & c8, row-wise prompt, answer temperature 0.3, program transfer, 96K 1-input / 64K multi, early second
attempts, $T=0.6$ & 53.33 \\
296 & first attempt on three inputs together, second attempt revises earlier answers & 52.50 \\
296 & as the first run with the row-wise prompt, with two grids per answer, second attempt on disagreement, vote,
batch-sharded sampling, 12 h 15 min & 49.86 \\
296 & as the first run with the row-wise prompt, with batch-sharded sampling and draft option, 12 h 15 min & no score \\
296 & tool agent for attempt 1, batch-sharded sampling, draft option, 11 h 50 min & 54.44 \\
296 & same configuration, second draw & 55.28 \\
296 & 75K agent budget for 1-input tasks (caps 59,000 / 8,000) & \textbf{56.67} \\
\bottomrule
\end{tabularx}
\end{table*}

On the same hardware and under the same limit, the best \qflash{} run exceeds the best \qbig{} run by 12.64 points (56.67
against 44.03), more than twice the resolution of single draws. The comparison is between systems, not models: the two runs
differ in the inference procedure as well as in the model.

%% file: sec/results.tex
\section{Results}
\label{sec:results}

\paragraph{Measurement.} Unless noted, every number in this section is a semi-private score on the 120 hidden ARC-AGI-2
tasks. The published description of the protocol says that the score averages over all test outputs. The scores
returned for our runs follow a mean over tasks instead: each of the 89 distinct scores among them is a multiple of
100/720, which is what averaging per-task fractions over 120 tasks with one, two or three test inputs gives, while a mean
over $N$ test inputs gives multiples of $100/N$, and no $N$ from 100 to 799 other than 720 fits all of them. We
therefore use the task-weighted mean offline as well; one fully solved task is worth 0.83 points. Each run is one
stochastic twelve-hour run. Nine \nvarc{} configurations were run two to five times without change (27 draws); their
pooled within-configuration standard deviation is 1.94 points (95\% interval 1.46 to 2.87). This estimate holds for the
\nvarc{} pipeline. For \qbig{} and \qflash{} we have two repeated pairs (two identical \qbig{} configurations that both
scored 26.81, and two draws of the 54.44 \qflash{} configuration, 54.44 and 55.28), too few to estimate their variation.
Where we use 1.94 points for these models below, it is an assumption: under it, two configurations measured once each
must differ by more than about 5.4 points to be told apart, and the best of five draws of one configuration exceeds its
mean by about 2.3 points on average. We therefore report means over repeated draws where they exist, and treat single-draw
differences of the other two models as descriptive.

\paragraph{Data experiments.} \Cref{tab:data-summary} collects the comparisons of training and calibration data from
\cref{sec:nvarc,sec:qwen,sec:flash}. For the grid model, the model continued on real tasks is the best single model, and
the best system runs it on all four GPUs; no checkpoint trained with synthetic tasks, alone or next to it, raised the
mean score. For the reasoning model, no later training set beat
\rftv{1} at the same cap, and at a matched 64K cap \rftv{1} itself differs from the base model by 6 of 172 inputs
($p=0.49$). For the expert lists, the calibration data mattered at a fixed number of experts: at 288 experts, the list
chosen by ARC routing frequency alone reached 58 correct dev attempts, the saliency list calibrated on coding traffic 66,
and the saliency core with its least-used experts exchanged by ARC routing 80.

\begin{table*}[tp]
\centering
\caption{Data experiments across the three models. Each row compares a training set (\nvarc{}, \qbig{}) or the data
that choose the experts to keep (\qflash{}) with a control. The two sides share the evaluation; unless a row says ``same
code'' or ``same configuration'', or uses the dev panel, they can also differ in training length, initialisation or
runtime, so most rows compare configurations rather than isolate the data. \lb{}: semi-private score, mean $\pm$ standard
deviation over draws, with the number of draws in parentheses. Offline: test inputs solved of the 172 public evaluation
inputs by one attempt at the given cap, in the 1-request two-slot harness unless marked 2-request. Dev: correct attempts
of 246 on the dev panel at a 96K cap (82 inputs $\times$ three seeds). The answer-only control was measured by the ARC Prize
Foundation under its own protocol.}
\label{tab:data-summary}
\footnotesize
\begin{tabularx}{\linewidth}{@{}>{\raggedright\arraybackslash}X>{\raggedright\arraybackslash}p{2.5cm}r>{\raggedright\arraybackslash}p{3.1cm}@{}}
\toprule
Training or calibration data & Metric & Result & Control \\
\midrule
\multicolumn{4}{@{}l}{\emph{\nvarc{} grid model (\cref{tab:nvarc-lb}); control: the released model unless stated}} \\
1,920 real tasks, continued fine-tuning & \lb{} & 31.72 $\pm$ 1.52 (22) & 28.96 $\pm$ 0.66 (4) \\
\quad then 25\% FORGE50 episodes & \lb{} & 29.31 (1) & parent: 31.72 $\pm$ 1.52 (22) \\
86\% FORGE50 episodes, 12\% real and 2\% hand-made tasks & \lb{} & 28.47 (1) & 28.96 $\pm$ 0.66 (4) \\
1,920 real and 95,993 corpus tasks & \lb{} & 28.61 (1) & 28.96 $\pm$ 0.66 (4) \\
1,920 real and 5,580 synthetic tasks & \lb{} & 28.06 (1) & 28.96 $\pm$ 0.66 (4) \\
shape-balanced curriculum of 36,000 tasks & \lb{} & 26.88 $\pm$ 0.49 (2) & 28.96 $\pm$ 0.66 (4) \\
FORGE50 continuation in place of ckpt-34000 on one of four GPUs, same code & \lb{} & 32.64 $\pm$ 1.91 (3) & ckpt-34000 on four GPUs: 33.06 $\pm$ 1.57 (5) \\
\midrule
\multicolumn{4}{@{}l}{\emph{\qbig{} reasoning model (\cref{tab:qwen-data,tab:qwen-models}); control as stated}} \\
answer-only, 37,201 examples, thinking off & \lb{} & 2.08 (1) & base, thinking off: 1.5\% \\
\rftv{1}: 1,036 verified self-traces & offline, 64K, 2-request & 63 & base: 57 \\
\rftv{2}: $+$777 traces, 755 with program targets & offline, 32K & 49 & \rftv{1}: 51 \\
\rftv{3}: $+$hard sampling & offline, 32K & 42 & \rftv{1}: 51 \\
short16k: traces of at most 16K tokens & offline, 32K, 2-request & 34 & \rftv{1}: 37 \\
v3b: 1,000 synthetic explanations in the thinking & offline, 32K, 2-request & 9 & \rftv{1}: 37 \\
ft64k: 1,584 synthetic explanations as answer text & offline, 32K & 43 & \rftv{1}: 51 \\
FINAL: $+$452 traces of other models & \lb{}, same configuration & 29.03 (1) & \rftv{1}: 44.03 (1) \\
\midrule
\multicolumn{4}{@{}l}{\emph{\qflash{} expert lists (\cref{tab:flash-prune}); control as stated}} \\
REAP-384: saliency on coding traffic & dev & 87 & unpruned: 78 \\
ARC-288: routing frequency on ARC training tasks & dev & 58 & REAP-288: 66 \\
s288p: REAP-288 core, least-used experts exchanged by ARC routing & dev & 80 & REAP-288: 66 \\
c8: s288p $+$8 experts per layer for program tokens & programs that do not compile, matched replay & 5.5\% (9/165) & s288p: 37.6\% (62/165) \\
\bottomrule
\end{tabularx}
\end{table*}

\paragraph{Runs by family.} We made 189 runs on the semi-private set. \Cref{tab:families} groups all of them. The 136 runs of the \nvarc{}
pipeline with our checkpoints averaged 29.57; among configurations with repeated draws, the best averaged 33.06 over five draws
(one single-draw configuration scored 33.89). The \qbig{} configurations without our fine-tuning averaged 22.66 and reached 26.81
at best; the fine-tuned \qbig{} line averaged 32.35 (34.44 without two runs that answered only half of the tasks by design)
and reached 44.03. Runs that combined both models scored at most 40.28. The ten scored runs of the
pruned and unpruned \qflash{} on the Merlin engine averaged 46.49 and reached 56.67.

\begin{table*}[t]
\centering
\caption{All runs on the semi-private set, grouped by method. The family rows count runs that scored 10 or more; the
last two rows count, across families, runs that returned no score or 0 (one of them a \qflash{} run that exceeded the time
limit) and completed runs that scored below 10 (five \nvarc{} runs, three \qbig{} runs including the answer-only
fine-tune at 2.08, and the two \qflash{} runs at effort \texttt{low} with stock vLLM, 8.61 and 6.11), so the family
means are conditional on a score of at least 10. ``Without our fine-tuning'' includes one third-party fine-tune of the
same base model.}
\label{tab:families}
\footnotesize
\begin{tabular}{@{}lrrrrr@{}}
\toprule
Family & Runs & Mean & SD & Median & Best \\
\midrule
\nvarc{} pipeline with the base checkpoint & 7 & 26.59 & 3.46 & 28.47 & 29.86 \\
\quad the released pipeline, unchanged & 3 & 29.12 & 0.70 & 29.03 & 29.86 \\
\nvarc{} pipeline with our checkpoints & 136 & 29.57 & 2.77 & 29.44 & 34.86 \\
\qbig{}, without our fine-tuning & 6 & 22.66 & 3.51 & 22.09 & 26.81 \\
\qbig{}, fine-tuned by us & 11 & 32.35 & 6.90 & 33.61 & 44.03 \\
\nvarc{} and \qbig{} in one run & 3 & 27.41 & 15.35 & 31.53 & 40.28 \\
\qflash{} on Merlin, 512 to 296 experts (\cref{sec:flash}) & 10 & 46.49 & 10.69 & 51.18 & 56.67 \\
Runs with no score or a score of 0 & 6 & -- & -- & -- & -- \\
Completed runs scoring below 10 & 10 & -- & -- & -- & -- \\
\bottomrule
\end{tabular}
\end{table*}

\paragraph{What moved the \qbig{} score.} \Cref{tab:qwen-steps} lists selected runs of the \qbig{} line in order,
with the factor that changed. Two steps changed the model. Replacing the base model by \rftv{1} also changed the cap from
64K to 32K and the temperature from 1.0 to 0.6, and the score rose from 26.81 to 28.19, within the variation of one draw; at
a matched 64K cap the two models differ by 6 of 172 inputs offline ($p=0.49$, \cref{sec:qwen-eval}). Replacing \rftv{1} by
FINAL in an identical configuration lowered the score from 44.03 to 29.03. With \rftv{1} held fixed, inference changes raised the
score from 28.19 to 44.03 in five single-draw steps: one request per attempt with the cap enforced in the engine and two
decode speed-ups ($+5.42$), the program in the same request ($+3.61$), NVFP4 with repair passes ($-0.97$), a 48K first
attempt ($+0.83$), and the 56K/32K schedule with deadline-aware caps ($+6.95$). Under the variation assumed above, only
the first and the last step reach the resolution of about 5.4 points on their own, the first by a small margin; their
sum, 15.84 points, is about three times that resolution.

\paragraph{Comparison with a measurement without a time limit.} The ARC Prize Foundation evaluated the base \qbig{} at
\texttt{xhigh} on the same 120 semi-private tasks without the twelve-hour budget, and reports 42.36 (305 of 720 points) at \$0.45 per task; at the
listed price of \$2.55 per million output tokens, this allows at most 175K output tokens per task
\citep{arcprizeleaderboard}. Our best run scored 44.03 (317 of 720) on these tasks in twelve hours on four L4 GPUs, with at
most 88K thinking tokens per test input over its two attempts. The difference is 1.67 points, or two fully solved tasks.
A difference this small cannot be resolved by single runs, but that does not show the two systems to be equal, and our
44.03 is the best of our runs rather than a mean; we therefore say only that our best twelve-hour run scored at the level
of the Foundation's measurement without the budget. On the public evaluation set both measurements use the same base model. The Foundation's per-task results
mark 40 of the 120 tasks as passed at \texttt{xhigh}; the base model in our serving setup (FP8 weights, the end-of-turn
fix, two attempts with 64K caps; \cref{sec:qwen-fixes}) answered every test input of 52 tasks (task score 49.86). We did
not rerun their protocol, so we cannot attribute the difference of 12 tasks to any single change.

\begin{table*}[t]
\centering
\caption{Selected runs of the \qbig{} line and the factor that changed at each step. Runs from 26.81 onward use four
L4 GPUs, tensor parallelism 4, MTP and an FP8 key-value cache. Rows below the rule are comparisons against the 44.03 run.}
\label{tab:qwen-steps}
\footnotesize
\begin{tabularx}{\linewidth}{@{}l>{\raggedright\arraybackslash}Xr@{}}
\toprule
Model & Configuration (change from the previous row) & \lb{} \\
\midrule
answer-only fine-tune & thinking off, 8 dihedral views, vote & 2.08 \\
base & thinking at low effort, end-of-turn masked & 9.86 \\
base & \texttt{xhigh}, 64K cap on both attempts & 26.81 \\
base & + program in a second request, $T=0.6$, reordered second attempts & 21.39 \\
base & 88K cap on both attempts & 18.89 \\
\rftv{1} & 32K cap on both attempts, $T=0.6$, two requests per attempt & 28.19 \\
\rftv{1} & one request, cap enforced in the engine, two decode speed-ups & 33.61 \\
\rftv{1} & + program in the same request & 37.22 \\
\rftv{1}, NVFP4 & + repair and extra passes & 36.25 \\
\rftv{1}, NVFP4 & first attempt 48K, second 32K & 37.08 \\
\rftv{1}, NVFP4 & first attempt 56K for every input first, second 32K on reordered demonstrations, deadline-aware caps & \textbf{44.03} \\
\midrule
FINAL, NVFP4 & same configuration as 44.03, model replaced & 29.03 \\
\rftv{1}, NVFP4 & task-weighted value queue (its check run took 68 minutes to load the model; cause of the drop unresolved) & 38.19 \\
\rftv{1} + \nvarc{} & 4.5 h of \nvarc{}, then the 44.03 schedule, then a merge & 40.28 \\
\bottomrule
\end{tabularx}
\end{table*}

%% file: sec/discussion.tex
\section{Discussion}
\label{sec:discussion}

\paragraph{What the data experiments show.} For the grid model, real tasks were the most useful training data that we
tested: continued fine-tuning on the 1,920 real tasks raised the mean from 28.96 to 31.72, and every model trained with
synthetic tasks that we ran alone scored below it (29.31, 28.61, 28.47, 28.06, 27.22 and 26.53; \cref{tab:data-summary}).
The rule corpus did not raise the score in our runs: a checkpoint trained with all 95,993 of its tasks scored 28.61 alone,
and checkpoints trained on it did not raise the mean when they ran next to the real-task model. Its tasks differ from
ARC-AGI-2 in grid size and in the absence of fixed colour meaning (\cref{sec:rules-compare}), which may limit transfer to
real tasks; we did not test this. For the reasoning model, the training data changed
less than the inference procedure: at a matched 64K cap, \rftv{1} and the base model differ by 6 of 172 inputs offline
($p=0.49$), and none of the later sets with synthetic tasks beat \rftv{1} (\cref{tab:qwen-synth}). Fine-tuning on the
model's own answer-verified traces did not lengthen the thinking in our runs: mean thinking at a 64K cap was 51.2K tokens
for \rftv{1} and 51.8K for the base model, although the selection by correctness and length changes the distribution of
the traces. A set with 452 traces from other models closed later (20 natural completions between 56K and 64K tokens
against 9) and scored 29.03 in the configuration in which \rftv{1} scored 44.03; FINAL also changed other parts of the
data, so a cost of off-policy traces under a fixed cap remains a hypothesis that we did not test. For
pruning, the calibration data decides which experts stay. At the tested concurrency the decode step of the L4 GPUs
changed little with the number of experts (about 258 to 264 milliseconds per cycle at 56 requests for 224 to 288
experts), so the main observed benefit of removing experts was memory, and in our tests the experts chosen by their
contribution to the output (saliency) worked better than those chosen by how often ARC tokens use them: the coding-calibrated saliency lists and the ARC routing-frequency lists selected substantially
different experts, and the frequency list completed its thinking early and lost accuracy (\cref{sec:flash-after}). Routing statistics from ARC training tasks served to exchange the least-used experts of
a saliency list, not to rank the experts on their own.

\paragraph{Why the inference changes work.} Within the twelve-hour budget, the tokens that a reasoning model can spend on
each hidden input constrain its score, and three measurements point this way. First, the same models score far higher without the
limit: the base \qbig{} reaches task scores of 49.86 to 54.03 on the public evaluation set with two long attempts, but 18.89
to 26.81 on the semi-private set with similar caps, while the schedule that fits the twelve hours scores 44.5 offline and
44.03 on the semi-private set (\cref{tab:qwen-val}). Second, naturally completed attempts were more accurate in these runs: attempts
whose thinking was closed at the token cap were right in 4\% to 9\% of cases for \qbig{} and in 5 of 46 cases for
\qflash{}, against 42\% to 86\% for natural completions (\cref{tab:qwen-length}). Third, the unpruned \qflash{} at 342 tokens per second leaves
an estimated 160 of about 343 hidden inputs without any answer at 96K caps, and the pruned model 46. The changes that
helped turned the fixed budget into more natural completions: one request per attempt with the cap enforced in the engine
(38\% more completed attempts per hour in a 40-task test, with the same 20 of 59 inputs solved), a schedule that starts
every first attempt before any second one, with deadline-aware caps ($+6.95$ points),
and a larger key-value cache through pruning (cache $\times$2.5, throughput $+52\%$). A program checked on the demonstrations
adds a second answer channel of high precision for few tokens: 73\% to 78\% of train-perfect programs were right on the test
input, at 4.5\% of the generated tokens, and the score rose from 33.61 to 37.22 when programs were added.

\paragraph{What may carry over beyond ARC.} Several components could be reused beyond ARC, given appropriate
calibration data and a task-specific verifier; the rule language, the palette key, the shortcut solvers, the dihedral
transforms and the output grammar are specific to grids, and we have not evaluated transfer to other domains. Enforcing a
thinking budget inside the serving engine and ordering attempts by a deadline could serve a reasoning model that must
answer many questions under a fixed compute budget. Pruning experts by saliency, adapted with routing on the target domain
and with experts kept for an output format such as code, could serve a mixture-of-experts model on GPUs with too little
memory. Rejection-sampling fine-tuning needs an exact checker of the final answer, and checking a program against the
given examples needs a few-shot problem with executable examples. The admission tests of the rule corpus need a program
language with an executable semantics and a finite set of probes and competitors.

\paragraph{What is new.} Relative to public work, we contribute (i) a synthetic corpus whose rules are checked to have
distinct behaviour signatures and whose tasks are checked against a finite set of competing rules, checks that the
descriptions of re-arc, BARC and NVARC do not report (\cref{tab:gen-compare}); (ii) ARC-adapted expert pruning that keeps the core of a saliency list, exchanges only its
least-used experts by task routing, and protects experts for program tokens, together with the observation that
coding-calibrated saliency lists and ARC routing-frequency lists select substantially different experts and that frequency
alone performed worse in our tests; (iii) a single-request attempt in which the
engine closes the thinking at the cap and a grammar constrains only the answer, with its measured effect on attempts per
hour; and (iv) a comparison of offline validation without a time limit with the semi-private score, which suggests that
throughput and token allocation constrain the deployed score, without isolating how much of the gap they cause. Answer
checking of self-generated traces, expert pruning by saliency and checking programs on demonstrations build on prior work
(\cref{sec:background}).

\paragraph{Remaining headroom.} Our best \qbig{} run scored at the level of the base model without a time limit,
and the fine-tuned model solves 99 of the 172 public evaluation inputs (57.6\% of inputs, task score 59.6) in at least
one of 8 to 22 attempts, against 71.5 inputs (task score 44.5, a mean over four pairings of attempts) in a two-attempt
schedule. Within this budget, further gains must come from more tokens per input or from more
accuracy per token. The \qflash{} line shows that the first route is open on the same hardware: with the same kind of
inference procedure, pruning and serving changes took a model of the same family from 25.00 to 56.67. The appendices list
unsuccessful variants and their measured results.

\paragraph{Experimental record.} The appendices give the hyper-parameters and data provenance (\cref{app:hparams}), the
run-to-run noise of repeated runs (\cref{app:noise}), and the unsuccessful and incomplete experiments (\cref{app:failed}),
including 16 expert lists and checkpoints of \qflash{} that we rejected (\cref{tab:flash-failed}); the synthetic training
sets of \qbig{} that did not beat \rftv{1} are in \cref{tab:qwen-synth}.

%% file: sec/conclusion.tex
\section{Conclusion}
\label{sec:conclusion}

We compared training and calibration data for three open models on ARC-AGI-2 under a budget of twelve hours on four L4
GPUs: a 4B grid model, a 27B reasoning model and a mixture-of-experts reasoning model. For the grid model, we built a
corpus of 12,000 rules with distinct behaviour signatures, with 95,993 tasks that were checked against a shortlist of
competing rules; continued training on 1,920 real tasks, run on all four GPUs with
revised scheduling, raised the semi-private score from 28.96 to 33.06 (mean of five runs, best 34.86), while no checkpoint
trained with synthetic tasks, including one trained on all 95,993 corpus tasks, improved on the real-task model. For the
27B reasoning model, fine-tuning on 1,036 answer-verified self-traces together with an inference
procedure built around the model (the thinking cap enforced in the engine, a program checked on the demonstrations, a long
first attempt for every input) raised the score from 26.81 to 44.03, the level of the ARC Prize Foundation's measurement of the base
model without the time limit (42.36); offline validation of the deployed schedule agreed with the semi-private score, and
the comparison suggests that throughput and token allocation constrain the deployed score. For the mixture-of-experts model, memory rather than compute limited the L4 GPUs:
removing a quarter of the experts by router-weighted activation saliency kept accuracy on our dev panel and grew the
key-value cache 2.5 times, lists adapted with ARC routing from training tasks and with experts kept for program tokens
reached 296 experts and, with an FP8 indexer cache, 3.6 times the cache, and the score rose from 25.00 to 56.67.

Our measurements have limits. All hidden-set scores are semi-private scores; the private scores of the same runs were not
available to us. Each run is one draw with a standard deviation of about 1.9 points, measured on repeated \nvarc{} runs
and assumed for the other two models, so single-factor comparisons below about 5.4 points are not resolved, and several
of our steps changed more than one factor. The \nvarc{} base model, ckpt-34000 and its descendants, and every \qbig{} training set from \rftv{1} on contain
public evaluation tasks, so offline numbers on those tasks are in-sample for the affected models; the subset H is
training-disjoint only for the \qbig{} line before FINAL, and we used it repeatedly during development, so it is a
development subset rather than an untouched test set. Our best \nvarc{} configuration changed the context lengths and the
scheduling together, so we cannot say how much each change contributed. Only one \qflash{} configuration has been scored twice (54.44 and 55.28), its accuracy gates ran
on a different engine and checkpoint from those used for hidden evaluation, and its expert lists were chosen by accuracy on public evaluation
panels. \Cref{app:failed} lists unsuccessful variants and their measured results. A direction we did not try is to measure
transfer to real tasks while a synthetic corpus is being built, and to admit rules by their effect on a held-out set of
real tasks as well as by their distinctness.

\section*{Acknowledgements}
We thank the authors of NVARC and of the Qwen models for releasing their weights and code, and the ARC Prize Foundation for
ARC-AGI-2 and its evaluation of hidden tasks. Development runs used Google Colab, rented GPUs and machines with four L4
GPUs.

%% file: sec/appendix.tex
\section{Hyper-parameters and setup}
\label{app:hparams}

\paragraph{\nvarc{} test-time training.} Fresh LoRA adapter per task, rank 256, $\alpha=32$, rank-stabilised, dropout 0,
on the query, key, value, output, gate, up and down projections plus the embedding and output layers (trained in full).
Data: 128 sequences (8 dihedral transforms $\times$ 16 random colour permutations of all ten colours, demonstrations
shuffled and cut from the front to fit the context). Loss on every output grid and the end-of-sequence token. One epoch,
batch 1 (128 steps), learning rate $5\times10^{-5}$, 10\% warm-up, cosine decay, gradient clipping 1.0, no weight decay.
Inference: 16 views per test input in 4 batches of 4, depth-first search with a cumulative-probability cut-off of 0.2 and
540 seconds per batch, rescoring under 8 views (one colour permutation each), selection by search hits minus mean rescoring
negative log-likelihood.

\paragraph{ckpt-34000.} Initialised from \nvarc{}; 1,920 tasks (ARC-AGI-1 training and evaluation, ARC-AGI-2 training,
ARC-AGI-2 public evaluation with test pairs appended), 16 views per task, 30,720 examples. Rank-256 rank-stabilised LoRA
with $\alpha=32$ on the same targets, learning rate $5\times10^{-5}$, 5\% warm-up, cosine decay over 61,440 planned steps,
batch 1, AdamW, context 8,192, loss on all output turns, checkpoint every 2,000 steps. One L4, stopped at step 34,273
after 10.0 hours. Software: Unsloth 2025.9.7, transformers 4.55.4, PyTorch 2.8.0.

\paragraph{\qbig{} \rftv{1}.} LoRA rank 32, $\alpha=32$, dropout 0 (233.5M parameters, 496 modules). AdamW (0.9, 0.999),
no weight decay, learning rate $10^{-4}$, 5\% warm-up, cosine decay, gradient clipping 1.0, one epoch, micro-batch 1
without packing, gradient accumulation 8, maximum length 81,920 tokens. Loss: token mean of cross-entropy over the
reasoning, the closing tag, the answer grid, the program (when present) and the end-of-turn token. Memory: re-entrant
gradient checkpointing, activation offload to host memory, MLPs chunked over 8,192 positions, chunked cross-entropy over
2,048 positions; peak 85~GiB on one RTX~PRO~6000 (96~GB). Software: transformers 5.17, PEFT 0.21.1, PyTorch 2.11. \rftv{1}:
1,033 rows, 130 steps, 7.06 hours. FINAL: 1,801 rows, 226 steps, 14.6 hours, maximum length 65,536.

\paragraph{Trace generation.} vLLM 0.30, maximum model length 81,920, 40 sequences, 32 concurrent requests; $T=0.6$,
top-$p=0.95$, top-$k=20$, \texttt{xhigh}; thinking cap 48,000 tokens, answer at most 8,192 tokens under a grid grammar;
seed derived from the task, test index, attempt and run tag. One generation pass over 490 training inputs took 5.2 hours
and produced 19.4M completion tokens.

\paragraph{\qbig{} on four L4 GPUs (44.03 run).} vLLM 0.30, tensor parallelism 4 on four L4 GPUs, NVFP4 weights (W4A16 Marlin
kernels on L4), FP8 key-value cache, three-token MTP with the end-of-turn rows of the output head zeroed, batch-sharded
sampling, maximum model length 98,304, 36 sequences, 32 concurrent client requests, 4,096 batched tokens. Caps: 56,000
thinking tokens for attempt 1 and 32,000 for attempt 2; deadline-aware caps at 15 tokens per second per stream; new
requests stop 15 minutes and the run stops 7 minutes before the 12-hour limit.

\paragraph{\qflash{} on four L4 GPUs (54.44 run).} Merlin 3.0 (vLLM 0.32.0, PyTorch 2.13, CUDA 13.0) with its profile for four
24~GB Ada GPUs: tensor parallelism 2 and pipeline parallelism 2 with expert parallelism, stages of 25 and 23 layers, custom
all-reduce off. The uploaded REAP-384 checkpoint is pruned at start-up to the 296 experts of c8 in 49 mixture-of-experts
layers (38,808 tensors dropped). FP8 key-value cache with static scales and an FP8 indexer cache, 7.7~GiB reserved per GPU
(1,957,095 tokens); maximum model length 114,688; 2,048 batched tokens; no prefix caching; a server limit of 64 sequences and 56 concurrent client requests; four
draft tokens with probabilistic draft sampling and an INT8 draft head; batch-sharded sampling; effort \texttt{xhigh},
$T=0.6$ for thinking, answer logits multiplied by 2 after the closing tag. Prompts state the grid sizes and give one row
per line. Attempt 1 is the tool agent with at most two checks and a total of 96K tokens for single-input tasks (80K for the
first turn, 8K for later turns) or 64K for multi-input tasks; attempt 2 has a 48K cap on reordered demonstrations and is
queued when attempt 1 finishes, ordered by a confidence class and then by thinking length, with an early second attempt for
single-input tasks as described in \cref{sec:flash-lb} (coverage guard off); a 16K repair turn with per-pair program
feedback is queued after attempt 1 only if no program is train-perfect and a plain-view program reproduces at least one
demonstration, and starts at least 45 minutes before the hard stop; one extra 48K attempt runs only if no program is
train-perfect, the two grids differ and at least 50 minutes remain. Deadline-aware caps assume 8 tokens per
second per stream and also cut the total budget of the agent, and a watchdog stops the run 11 hours and 50 minutes after it starts. The 56.67 run differs only in the agent budget of
single-input tasks: 75K tokens in total, with caps of 59,000 and 8,000.

\paragraph{Data provenance.} The 30 H tasks of the public evaluation set were excluded from every training set of the RFT
line before FINAL; the answer-only fine-tune held out a different set of 30 tasks, which shares 8 tasks with H. ckpt-34000
and all \nvarc{} checkpoints derived from it were trained on all 120 public evaluation tasks. No hidden task was available
to us at any time; all semi-private numbers are on hidden tasks.

\section{Unsuccessful and incomplete experiments}
\label{app:failed}

For completeness, we list each idea that did not help, with its measured result, and in \cref{tab:flash-failed} every
expert list and smaller checkpoint of \qflash{} that we rejected. Together with the experiment tables of the main text,
this is the full record of what we tried for the three lines.

\paragraph{Answer-only fine-tuning of \qbig{}.} Training on 37,201 answer grids with thinking disabled gave 2.08 on the
semi-private set. A continuation on all 1,147 labelled official tasks solved 0 of 9 inputs of tasks in its own training data
(\cref{sec:qwen-traces}).

\paragraph{Explanations inside the thinking channel.} Placing 1,284 verified explanations (284 written by an LLM for official
tasks, 1,000 generated from templates for synthetic tasks) in the reasoning lowered accuracy at 32K from 37 to 9 of 172
inputs; most attempts opened with a solution summary (\cref{sec:qwen-eval}). Placing 1,602 explanations, 1,584 of them of
synthetic tasks, in the answer channel (ft64k) gave 43 against 51 (two-slot, production harness; \cref{tab:qwen-synth}).

\paragraph{More or shorter self-traces.} Adding 73 traces from an 88K-cap second pass (main2) gave 57 of 172 inputs at 64K,
equal to the base model. Training only on traces of at most 16K tokens gave 34 against 37 at 32K and did not shorten
thinking on hard tasks (83\% of attempts still reached the cap).

\paragraph{Program targets in the fine-tuning data.} \rftv{2} (1,655 program targets) and \rftv{3} (1,941, including 319
programs from other attempts) did not raise the rate of train-perfect programs (41 and 30 against 42 at 32K).

\paragraph{A larger final training set.} FINAL, with 452 teacher traces, 315 symmetry-view traces and all public evaluation
tasks, scored 29.03 against 44.03 for \rftv{1} in the same configuration.

\paragraph{Fine-tuning a program writer.} Qwen3-8B trained on 512 to 10,000 compiled programs never produced a program that
passes the demonstrations of any of 15 public evaluation tasks (0 of 120 programs). With 512 and 3,000 targets that
call the corpus library, 85 and all 120 of the programs failed while reading the palette key, which real tasks do not
have; 10,000 standalone targets without the library gave more malformed and truncated programs than the base model.

\paragraph{Synthetic curricula for \nvarc{}.} 36,000 shape-balanced tasks raised accuracy on new tasks of their generator
from 5.8\% to 61.7\% but scored 26.53 on the semi-private set. Two FORGE50 recipes scored below ckpt-34000 alone (mean 31.72): continuing
ckpt-34000 with 25\% FORGE50 episodes next to all 1,920 real tasks in 16 views scored 29.31, and a mixture of 86\%
FORGE50 episodes with 12\% real tasks and 2\% hand-made tasks scored 28.47. A checkpoint trained from the base model on
the 1,920 real tasks and all 95,993 rule-corpus tasks scored 28.61 alone.

\paragraph{A likelihood-threshold training loss.} A gate loss matched to the 0.2 search cut-off solved 2 of 18 inputs, the
same as the base model and the cross-entropy control.

\paragraph{Program retrieval.} None of the 11,997 library programs for tasks outside the public evaluation set reproduces
all demonstrations of any public evaluation task; three retrieved programs in the prompt gave 48 against 54 of 328
attempts.

\paragraph{Code agents.} A tool-calling agent with three code executions per attempt at 32K solved 5 of 20 inputs against 7
and 4 for direct answers in two seeds; 57 of its 82 calls were exploration, and 13 of the 20 inputs failed under every
protocol. Two runs of tool-integrated reasoning were invalid for different reasons: in the first (5 against 20 of 90) the
model ended its thinking to emit code and no code was executed; in the second (21 against 20) the sandbox's memory limit
broke 135 of 246 calls. We count tool-integrated reasoning as untested.

\paragraph{Combining \nvarc{} and \qbig{}.} Giving \nvarc{} the first 4.5 hours and \qbig{} the rest scored 40.28 against
44.03 for \qbig{} alone.

\paragraph{Allocation variants.} Uniform 88K caps scored 18.89; a program in a separate request scored 21.39 (with other
changes); a task-weighted value queue scored 38.19 (its check run needed 68 minutes to load the model); a soft wrap-up
sentence at 32K inside a 45K cap scored 26.39.

\paragraph{Expert lists for \qflash{}.} REAP-288 solved 66 and the ARC-frequency list ARC-288 58 of 246 dev attempts,
against 87 for REAP-384; CORE-320 failed the fresh-panel rule (lower bound $-4.46$ points). Cutting c8 to 272 or 256
experts lost 2.16 and 3.02 points on the production mix; at 224 experts we have only an L4 probe (17 right first
attempts of 32 against 21 for s288p with an FP8 indexer cache, $p=0.39$) and a thinking-only screen ($-6.25$ points), and
a fully protected 256-expert list lost 17.8 points with the agent. Teacher-forced likelihood ranked ARC-288 close to REAP-384 and did not predict its loss. \Cref{tab:flash-failed}
lists all 16 rejected lists and checkpoints.

\begin{table*}[t]
\centering
\caption{Expert lists and smaller checkpoints of \qflash{} that we rejected. ``Dev'' is correct attempts on the dev panel
at a 96K cap (246 with three seeds unless noted); ``production mix'' is the task-weighted score of thinking and programs as
used on the L4 GPUs, and ``thinking'' the score of the thinking alone, both as differences in points against c8. The adopted
lists were REAP-384, then s288p, then c8 (\cref{tab:flash-prune}).}
\label{tab:flash-failed}
\footnotesize
\begin{tabularx}{\linewidth}{@{}l>{\raggedright\arraybackslash}p{0.27\linewidth}>{\raggedright\arraybackslash}X>{\raggedright\arraybackslash}p{0.17\linewidth}@{}}
\toprule
List (experts) & Construction & Measured result & Why rejected \\
\midrule
REAP-448 (448) & public REAP list & dev 26 of 82 (one seed) against 30 unpruned; CUDA-graph capture out of memory at a 4.0~GiB
cache pin & frees less memory, not more accurate \\
REAP-288 (288) & public REAP list & dev 66 against 78 unpruned and 87 REAP-384 & accuracy \\
ARC-288 (288) & top 288 by ARC routing frequency $F_e$ & dev 58, lowest of all lists (23 gained, 43 lost against unpruned,
$p=0.019$); 85\% natural completions at 62.6K thinking tokens & accuracy \\
CORE-320 (320) & REAP-288 + 32 experts by $F_e$ & dev 84; fresh panel $+2.64$ points against REAP-384, lower bound
$-4.46$ & failed the adoption rule \\
CORE-300 (300) & REAP-288 + 12 experts by $F_e$ & dev 77 & accuracy \\
c336o (336) & CORE-320 + the 16 highest-$F_e$ experts that REAP-384 drops & dev 56 against 60 for REAP-384 (two seeds,
164); $+4.3\%$ throughput & speed gain below 5\% \\
s288p v2 (288) & s288p with exchanges chosen from the traces of lost tasks & no missing expert was used heavily on the 9
lost tasks & no target to fix; never run \\
c8$-$24 (272) & c8 minus its 24 lowest-$F_e$ experts outside the code experts and concentrated specialists & thinking $+0.83$; production
mix $-2.16$ & accuracy \\
c8$-$40 (256) & as above, 40 removed & thinking $+0.90$ over 344 pairs; production mix $-3.02$ (probability of a gain
0.05); cache $+14\%$ & accuracy \\
c8$-$72 (224) & as above, 72 removed & cache $+25\%$; L4 probe: 17 of 32 first attempts right, s288p 21 ($p=0.39$) &
accuracy \\
SAL-256 (256) & c8 cut by REAP saliency measured on ARC traces & thinking $-3.61$; production mix $-1.15$ & accuracy \\
FREQ-224 (224) & c8 cut by $F_e$ with soft protection & thinking $-6.25$ (probability of a gain 0.04) & accuracy \\
c8$-$40 protected (256) & variant of c8$-$40 with protected specialists and code experts & agent configuration $-17.8$ points
task-weighted, $-26.8$ on single-input tasks; median first-turn thinking 13,441 against 27,540 tokens & stops thinking early \\
e384 (384) & third-party list by routing share on ARC agent transcripts & not run & calibrated on 60 of the 120 public
evaluation tasks \\
REAM-288 (288) & third-party build with merged experts & not run & keeps the same experts as REAP-288 in all 48 layers \\
INT4 attention (512) & round-to-nearest INT4 for attention and Gated DeltaNet weights & teacher-forced log-likelihood
$-0.64$ nats per token; 29.9\% of tokens off by more than 1 nat & breaks the model \\
\bottomrule
\end{tabularx}
\end{table*}

\paragraph{Serving \qflash{}.} Stock vLLM fitted 94,001 cache tokens (four requests, 164 tokens per second); SGLang fitted
about 434,000 and generated 221 to 255 tokens per second against 639 to 657 for Merlin; six draft tokens did not boot and
three were 1.2\% slower than four; prefix caching was slower; a larger indexer budget of 8,192 positions cost 5.5\% of the
cache for a gain of 2.0 points that did not reach our threshold.

\paragraph{Prompts and attempts for \qflash{}.} One request for all test inputs of a task lost 7.95 points offline, two
single-input tasks in one request 5.7, rendered grid images 4.52, and code-first prompts 1.41 and 9.04. Effort
\texttt{medium} lowered the task-weighted score from 28.10 to 5.88, and $T=1.0$ closed fewer attempts by 64K (61 against
90 of 243) and solved fewer by 96K (65 against 78); its run scored 40.97. Showing the first-attempt grid and asking
for a fix broke 8 right answers and fixed 1 at a 2K budget. On the semi-private set, a first attempt on three inputs together
with a refining second attempt scored 52.50 and two grids per answer with voting 49.86, both below 53.33.

\twocolumn[{%
\section{Run-to-run noise}
\label{app:noise}
\begin{center}
\captionof{table}{Configurations run more than once without change (semi-private set). All are \nvarc{} pipelines. The pooled
within-configuration standard deviation over the 27 draws is 1.94 (95\% interval 1.46 to 2.87, 18 degrees of freedom).}
\label{tab:noise}
\footnotesize
\begin{tabularx}{\linewidth}{@{}lrrr>{\raggedright\arraybackslash}X@{}}
\toprule
Configuration & Draws & Mean & SD & Scores \\
\midrule
ckpt-34000, single model & 5 & 32.05 & 2.55 & 33.47, 34.44, 27.78, 32.36, 32.22 \\
ckpt-34000 on four GPUs with runtime changes (\cref{sec:nvarc-ens}) & 5 & 33.06 & 1.57 & 33.61, 34.86, 33.06, 33.19, 30.56 \\
ckpt-34000 and the FORGE50 continuation, two GPUs each & 3 & 31.30 & 2.76 & 34.31, 30.69, 28.89 \\
as the runtime-changes row, FORGE50 continuation on the fourth GPU & 3 & 32.64 & 1.91 & 34.31, 33.06, 30.56 \\
ckpt-34000, single model, another runtime & 3 & 31.39 & 0.70 & 31.39, 32.08, 30.69 \\
four further configurations & 2 each & 28.12--32.29 & 0.59--1.87 & \\
\bottomrule
\end{tabularx}
\end{center}
\vspace{1ex}}]

Two runs of \qbig{} with identical code both scored 26.81. Two draws of the 54.44 \qflash{} configuration scored 54.44 and 55.28.